\documentclass[letterpaper, 10 pt, conference]{ieeeconf}  % Comment this line out if you need a4paper

\IEEEoverridecommandlockouts                              % This command is only needed if 
\usepackage{graphicx}
\usepackage{import}
\usepackage{graphics}		% for pdf, bitmapped graphics files
\usepackage{epsfig}			% for postscript graphics files
\usepackage{graphicx}
\usepackage{bm}
\usepackage{amsmath}		% assumes amsmath package installed
\usepackage{amssymb}		% assumes amsmath package installed

\usepackage{algorithm}
\usepackage[noend]{algpseudocode}	% for IEEE-formatted pseudo-code usage in algorithms
\usepackage[table]{xcolor}
\usepackage{lipsum}
\usepackage{array}
\usepackage{float}
\usepackage{color}
\usepackage{booktabs}
\usepackage{multirow}

\usepackage{subcaption}
\graphicspath{ {fig/} }

\usepackage{xr}
\usepackage{soul}
\usepackage{textcomp}

\usepackage{siunitx}

\usepackage{epstopdf}
\usepackage{color}
\usepackage{bm,amsmath}

\usepackage{caption}
\usepackage{here}
\usepackage[normalem]{ulem}

\makeatletter
\let\NAT@parse\undefined
\makeatother
\usepackage[numbers,sort&compress]{natbib}
\usepackage[pagebackref=false,breaklinks=true,colorlinks,urlcolor=blue,citecolor=blue,linkcolor=blue,bookmarks=true]{hyperref}
\colorlet{light_blue}{blue!40}
\colorlet{light_green}{green!40}
\colorlet{light_red}{red!40}
\usepackage{tikz}
\usetikzlibrary{fit,calc}
\newcommand{\boxitt}[2]{
    \tikz[remember picture,overlay] \node (A) {};\ignorespaces
    \tikz[remember picture,overlay]{\node[yshift=3pt,fill=#1,opacity=.25,fit={($(A)+(-2.5,-0.9\baselineskip)$)($(A)+(0.3\columnwidth,-{#2}\baselineskip - 0.25\baselineskip)$)}] {};}\ignorespaces
}
\usepackage{titlesec}
\titlespacing{\section}{0pt}{0.4\baselineskip}{0.25\baselineskip}
\titlespacing{\subsection}{0pt}{0.25\baselineskip}{0.15\baselineskip}
\titlespacing{\subsubsection}{0pt}{0.05\baselineskip}{0.03\baselineskip}

\renewcommand{\paragraph}[1]{\vspace{0.1em}\noindent\textit{#1} --}
\usepackage{color,soul}

\definecolor{chartblue}{RGB}{21, 53, 98}

\newcolumntype{C}[1]{>{\centering\let\newline\\\arraybackslash\hspace{0pt}}m{#1}}

\algnewcommand{\IfThenElse}[3]{% \IfThenElse{<if>}{<then>}{<else>}
  \State \algorithmicif\ #1\ \algorithmicthen\ #2\ \algorithmicelse\ #3}

\algnewcommand{\IfThen}[2]{% \IfElse{<if>}{<then>}
  \State \algorithmicif\ #1\ \algorithmicthen\ #2}

\algrenewcommand\algorithmicrequire{\textbf{Input:}}
\algrenewcommand\algorithmicensure{\textbf{Output:}}

\title{\LARGE \bf
Chemistry Lab Automation via Constrained Task and Motion Planning
}
\author{Naruki Yoshikawa$^{\ast1}$,
        Andrew Zou Li$^{\ast1}$,
        Kourosh Darvish$^{\ast1,\dagger}$,
        Yuchi Zhao$^{\ast2}$,
        Haoping Xu$^{\ast1}$,\\
        Artur Kuramshin$^{1}$,
        Al\'{a}n Aspuru-Guzik$^{1}$,
        Animesh Garg$^{1,3}$,
        Florian Shkurti$^{1}$
\thanks{$^\ast$ Authors contributed equally
    $^{1}$University of Toronto \& Vector Institute
    $^{2}$University of Waterloo,
    $^{3}$Nvidia
    $^\dagger$corresponding author.}
}

\begin{document}
\bstctlcite{IEEEexample:BSTcontrol}

\maketitle
\thispagestyle{empty}
\pagestyle{empty}

%%%%%%%%%%%%%%%%%%%%%%%%%%%%%%%%%%%%%%%%%%%%%%%%%%%%%%%%%%%%%%%%%%%%%%%%%%%%%%%%

\begin{abstract}
% In the process of material discovery, researchers currently need to perform many laborious and often dangerous lab experiments. Robots in the chemistry laboratory are a scalable and autonomous tool that will enable accelerated material discovery and significantly reduce the time chemists spend on dangerous,  repetitive tasks. Current lab automation setups involve dedicated hardware for particular tasks, and any usage of robots require static, structured environments with significant investment or preparation beforehand. We present an adaptive robotic system that can load chemistry experiment descriptions, perceive a dynamic workspace, and autonomously plan the required actions and motions to perform the given chemistry experiments with common tools found in the existing lab environment. The proposed architecture utilizes integrated task and constrained motion planning with PDDLStream, enabling the system to generate plans that avoid objects in the workspace, reposition obstacles if necessary, and prevent chemical spillage while transporting vessels. As a result, the system can be safely integrated into existing experiment setups with minimal preparation.

Chemists need to perform many laborious and time-consuming experiments in the lab to discover and understand the properties of new materials. 
To support and accelerate this process, we propose a robot framework for manipulation that autonomously performs chemistry experiments.
Our framework receives high-level abstract descriptions of chemistry experiments, perceives the lab workspace, and autonomously plans multi-step actions and motions.
The robot interacts with a wide range of lab equipment and executes the generated plans.
A key component of our method is constrained task and motion planning using PDDLStream solvers. 
Preventing collisions and spillage is done by introducing a constrained motion planner.
Our planning framework can conduct different experiments employing implemented actions and lab tools.
We demonstrate the utility of our framework on pouring skills for various materials and two fundamental chemical experiments for materials synthesis: solubility and recrystallization\footnote{
More experiments and supplementary materials can be found at: \\\url{https://ac-rad.github.io/robot-chemist-tamp/}}.

 %and efficiently makes use of commonly available lab tools.
 %for integrated task and constrained motion planning 
%which generates task plans and constrained motion plans.
%Those plans are preventing collisions and spillage.
 %, without requiring the design of custom finite-state machines for each experiment.
%Our overall planning framework can scale to many different experiments, actions, and lab tools %, without requiring the design of custom finite-state machines for each experiment.

%We demonstrate the utility of our framework on pouring skills for various materials and two foundational chemical experiments for materials synthesis: solubility and recrystallization. More experiments, supplementary materials, and updated evaluations can be found at \textcolor{blue}{\url{https://ac-rad.github.io/arc-iros2023/}}. 

% that avoid objects in the workspace, reposition obstacles if necessary, and prevent chemical spillage while transporting vessels. As a result, the system can be safely integrated into existing experiment setups with minimal preparation.
% Our proposed approach can be extended to different scenarios.

% This work is towards accelerating material discovery using robots. In this work, a collaborative robot is used to load the chemistry experiment description, perceive the workspace, and plan its actions and motions to perform the given chemistry experiments. The proposed architecture is based on integrated task and constrained motion planning using PDDLstream to perform the experiment.
% \AniG{make a website and include results, videos and stuff...}
\end{abstract}


\section{Introduction}
\label{sec:Introduction}
% \begin{itemize}
%     \item maybe after lab automation description
%     \item highlighting safety or chemistry lab constraints requirement considerations  (involving task and motion planning)
%     \item highlighting chem-specific skills and their requirements: manipulation (pouring+ lab instruments) and perception (turbidity) skills
% \end{itemize}
Chemistry experiments are essential for finding novel materials or verifying hypotheses in materials science. These experiments are typically conducted by human chemists. They are laborious, time-consuming, and often challenging to reproduce. There is a growing effort to realize \emph{self-driving labs}~\cite{seifrid2022autonomous,abolhasani2023rise}
reducing the work of chemists and accelerating material discovery. These are automated chemistry labs that close the loop between autonomously (a) selecting the next experiments to run, (b) executing them in the lab using robotic equipment, and (c) evaluating their outcomes, feeding back the results to the experiment selection module. 

\textbf{Challenges and Proposed Solutions.}
One of the main challenges in realizing the promise of full automation in chemistry labs is the over-reliance on specialized hardware that is tailored to conduct a small part of the overall experiment, and is often immutable and not programmable. This makes adaptation of custom lab equipment to perform multiple types of experiments exceedingly difficult. 

Our work aims to address this challenge by enabling general-purpose robot manipulators to autonomously execute chemistry experiments that are described in a high-level abstract specification language. We present a robotic system that incorporates perception, constrained task and motion planning, and vision-based evaluation of experimental outcomes. It satisfies requirements (b) and (c) mentioned above for a number of fundamental chemistry experiments that involve pouring of liquids, solubility, and re-crystallization.  
%Moreover, current chemistry equipment is designed for human use in mind%, and laboratory environments are semi-structured. 

%Any general-purpose robotic solution for lab automation must operate in semi-structured environments and use tools designed for humans. Chemists, as end users, would need to interact with and fully leverage such robotic platforms to perform experiments.
%An approach is to facilitate robots by getting high-level experiment descriptions from chemists. The robot should be able to adapt to different workspace conditions while mapping chemistry descriptions to robot actions.
% , who typically do not have robotics background,
%while achieving the experiment goals.
% while accounting for safety considerations. 

\begin{figure}[t]
    \includegraphics[trim={0 1cm 0 4cm}, clip,width=0.8\columnwidth]{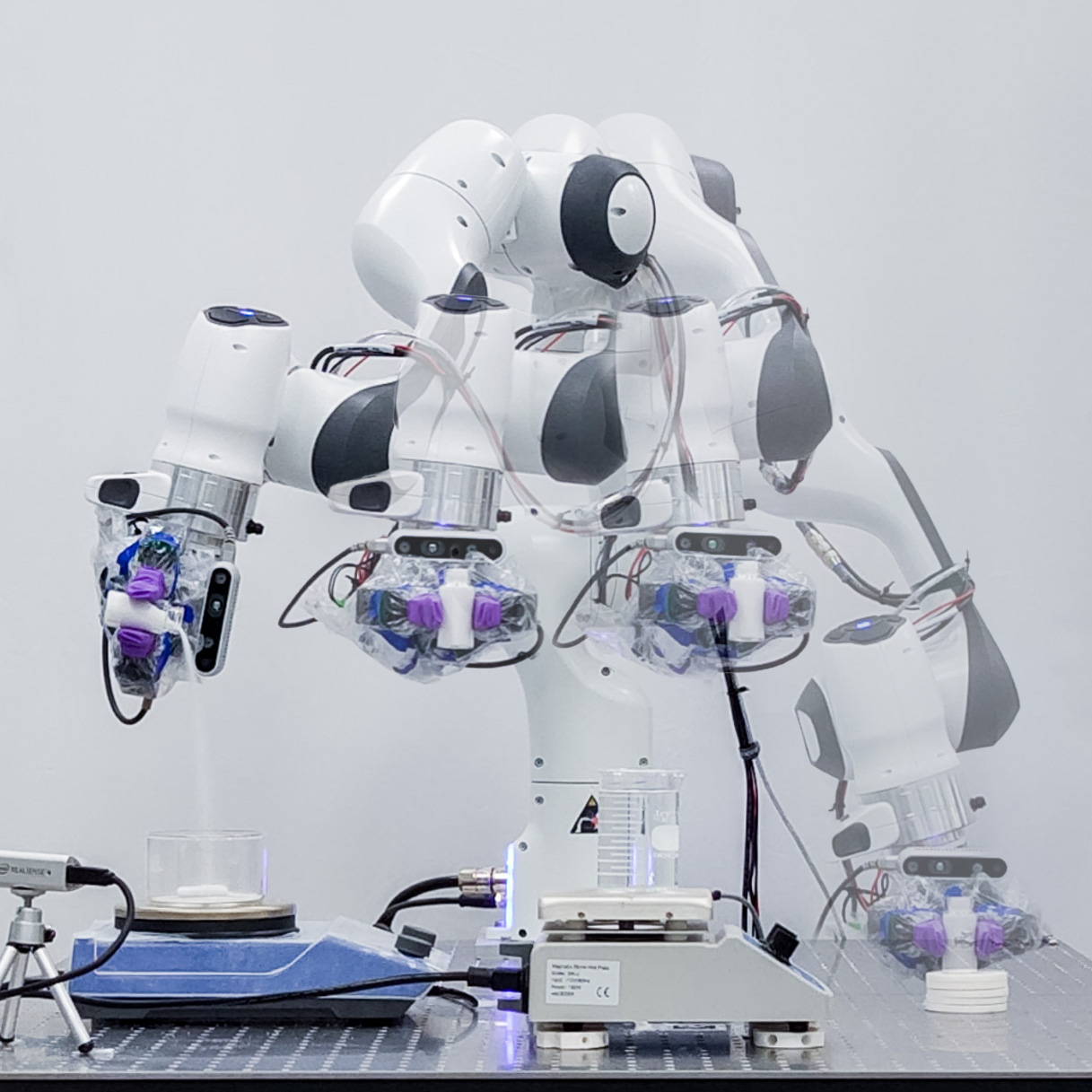}
    \centering
    \caption{Robot pouring granular material during an experiment.}
    \label{figure_1}
\end{figure}
A second central challenge in chemistry lab automation is ensuring safety during experiment execution and interactions with chemists~\cite{menard2020review}. Safety requirements can be multi-layered. At a high-level, in terms of experiment description and task planning, we can impose the constraint that materials can be synthesized in a specific order. At a low-level, in terms of manipulation and perception skills, robots should adjust their motion so as to not spill the contents of chemistry vials and beakers during transportation. Our system is able to effectively address both types of safety requirements by expressing them as constraints in terms of task planning as well as motion planning.

% While they should perceive objects in the workspace, read the object labels, plan grasp, and physically interact and manipulate with the objects, they should be able to perceive transparent objects, handle specialized lab tools, or perceive the chemistry experiment progress. 

%.
\textbf{Desiderata and Capabilities.} There are a number of key enabling capabilities that are needed to robustly address the challenges above. Chemistry robots are required to be equipped with a rich repertoire of general and chemistry domain-specific perception and manipulation skills. They must be able to recognize transparent objects such as glassware, opaque tools, liquids, powders and other substances; segment and recognize the contents of vessels; and estimate object poses~\cite{xu2021seeing, Wang2023MVTrans}. Robots should perceive and monitor the state of the material's synthesis; for example, they should detect when a solution is fully dissolved~\cite{shiri2021automated}. They need to perform dexterous manipulation and handling of objects. Some examples are constrained motion generation for picking and transporting beakers while avoiding spillage of their contents; pouring skills; manipulating tools, rigid objects, and deformable objects. Moreover, robot-executed chemistry experiments require high precision and repeatability to achieve reproducible and reliable results.

\textbf{Contributions.} Our paper presents three major contributions:
(I) An autonomous robotic system and modular system for chemistry lab automation that receives as input an abstract experiment description from chemists, perceives the environment, and plans long-horizon trajectories that perform a diverse set of multi-step chemistry experiments. This is an advancement from \cite{fakhruldeen2022archemist}, where a finite state machine with fixed objects in a static workspace was used to perform chemistry experiments.
(II) We incorporate constrained motion planning in a PDDLStream~\cite{ICAPS20paper186} solver. This enables long-horizon, multi-task planning and reasoning, while avoiding spillage when transporting liquids and powders. To enhance the success rate of planning in the presence of motion constraints, we have shown that the addition of a degree of freedom to commercial 7-DoF robot arms can be beneficial.
% (though not necessary).
We show that the 8-DoF robot has 97\% success rate in constrained motion planning compared to 84\% of the 7-DoF robot.
% We show the 8-DoF robot has XX.X\% success rate in constrained motion planning compared to that higher constrained motion planning success rate compared to unconstrained motion planning. 
(III) We present and analyze a set of accurate and efficient pouring skills inspired by human motions. It has an average relative error of 8.1\% and 8.8\% for pouring water (liquid) and salt (granular solids) compared to a baseline method with 81.4\% and 24.1\% errors. These results are comparable with recent results~\cite{Kennedy2019Autonomous, Huang2021Robot}, while our method is simpler, and require fewer and simpler sensors.

As a proof of concept, we have shown that by closing the loop with the perception of the current status of the chemistry experiment under execution, we can achieve results comparable to the literature ground truth for \textit{solubility} experiment. We attained 7.2 \% error for the solubility of salt and successfully recrystallized alum. 

% As a proof of concept in using robots for scientific finding in chemical synthesis domain, we have shown by closing the loop in chemistry experiment task execution, we could achieve results comparable to the literature ground truth for \textit{solubility} experiment.
% Solubility is the maximum amount of a substance (\textit{solute}) that can dissolve in a specific amount of another substance (\textit{solvent}) at a specific condition.
% We attained $3.42\%$ error for the solubility of salt and successfully recrystallized alum. 
% We validate the utility of our framework by conducting solubility measurement experiments. 

\section{Related Work}
\label{sec:related-work}

% \subsection{Lab Automation}
\paragraph{\textbf{Lab Automation}}
Lab automation aims to introduce automated hardware in a laboratory to improve the efficiency of scientific findings.
An example of lab automation is the usage of mobile robots for improving photocatalysts for hydrogen production from water~\cite{burger2020mobile}. 
Recently, an automated workflow that translates organic chemistry literature into a structured language called XDL was proposed~\cite{mehr2020universal}. ARChemist~\cite{fakhruldeen2022archemist}, a lab automation system, was developed to conduct experiments including solubility screening and crystallization without human intervention.
% based on XDL descriptions.
Although these major steps towards chemistry lab automation have been made, their dependence on predefined tasks and motion plans without constraint satisfaction guarantees limits their flexibility in new and dynamic workspaces. In those works, pick \& place was the primary task that the manipulators were carrying out. 
Those works were tested in hand-tuned and static environments to avoid occurrences of unsatisfied task constraints, such as chemical spills from vessels filled with liquid during transfer.
Our framework resolves these gaps through constraint satisfaction and scene-aware planning with a variety of skills.

\paragraph{\textbf{Task and Motion Planning with Constraints}}
Task and motion planning (TAMP) simultaneously determines the sequence of high-level symbolic actions, such as picking and placing, and low-level motions for the action, such as trajectory generation.
%Among early approaches for TAMP, aSyMov~\cite{cambon2009hybrid} and SMAP~\cite{plaku2010sampling} interleaved symbolic planners and motion planners. 
%Monte Carlo tree search~\cite{ren2021extended} and in-the-loop simulation for decision-making~\cite{Darvish2018Interleaved} have also been applied to TAMP problems.
%FFRob~\cite{garrett2018ffrob} leveraged symbolic planners by creating symbolic actions from a sample of continuous motions.
Another TAMP solver, PDDLStream~\cite{ICAPS20paper186}, extends PDDL~\cite{aeronautiques1998pddl}, a common language to describe a planning problem mainly targeting discrete actions and states, by introducing streams, a declarative procedure via sampling procedures. PDDLStream reduces a continuous problem to a finite PDDL problem and invokes a classical PDDL solver as a subroutine.
Since PDDLStream verifies the feasibility of action execution during planning time, it can inherently enhance safety
% in chemistry laboratories
by avoiding unfeasible plans or plans that may lead to unsafe situations.
Nonetheless, PDDLStream does not yet account for constraints in the planning process, for example avoiding material spillage from beakers during transportation, which impedes its deployment in real-world lab environments.
For this purpose, sampling-based motion constraint adherence \cite{doi:10.1177/0278364910396389} or model-based motion planning \cite{muchacho2022solution} are possible stream choices.
To overcome this shortcoming, our work extends PDDLStream with a projection-based sampling technique \cite{Kingston2019Exploring} to provide constraint satisfaction, completeness, and global optimality.

\paragraph{\textbf{Skills and Integration of Chemistry Lab Tools}}
In the process of lab automation, robots interact with tools and objects within the workspace and require a repertoire of many laboratory skills.
Some skills can be completed with existing heterogeneous instruments and sensors in chemistry labs, such as scales, stir plates, pH sensors, and heating instruments.
Other skills are currently done either manually by humans in the lab or with expensive special instruments.
In a self-driving lab, robots should acquire those skills by effectively using different sensory inputs to compute appropriate robot commands.
Pouring is a common skill in chemistry labs.
Recent work \cite{Kennedy2019Autonomous, Huang2021Robot} used vision and weight feedback to pour liquid with manipulators. \cite{Kennedy2019Autonomous} proposed optimal trajectory generation combined with system identification and model priors. To achieve milliliter accuracy in water pouring tasks with a variety of vessels at human-like speeds, \cite{Huang2021Robot} used self-supervised learning from human demonstrations.
In this work, we have reached similar results for pouring, using commercial scales that have delayed feedback.
Our approach is model-free, and it can pour granular solids as well. Granular solids have different dynamics from liquids, similar to the avalanche phenomenon.
%\textcolor{blue}{It outperforms \cite{Kennedy2019Autonomous} in terms of pouring efficiency.}
Lastly, while executing a chemistry experiment, the robot should possess perception skills to measure progress toward completing the task. For example, in solubility experiments, the robot should perceive when the solution is fully dissolved, and therefore stop pouring the solvent into the solution.
There are different ways to measure solubility. In our work, we use the turbidity measure \cite{shiri2021automated}, which is based on optical properties of light scattering and absorption by suspended sediment \cite{kitchener2017review}.
\section{Methods}
\label{sec:methods}

\begin{figure*}[t!]
    \includegraphics[width=0.9\textwidth]{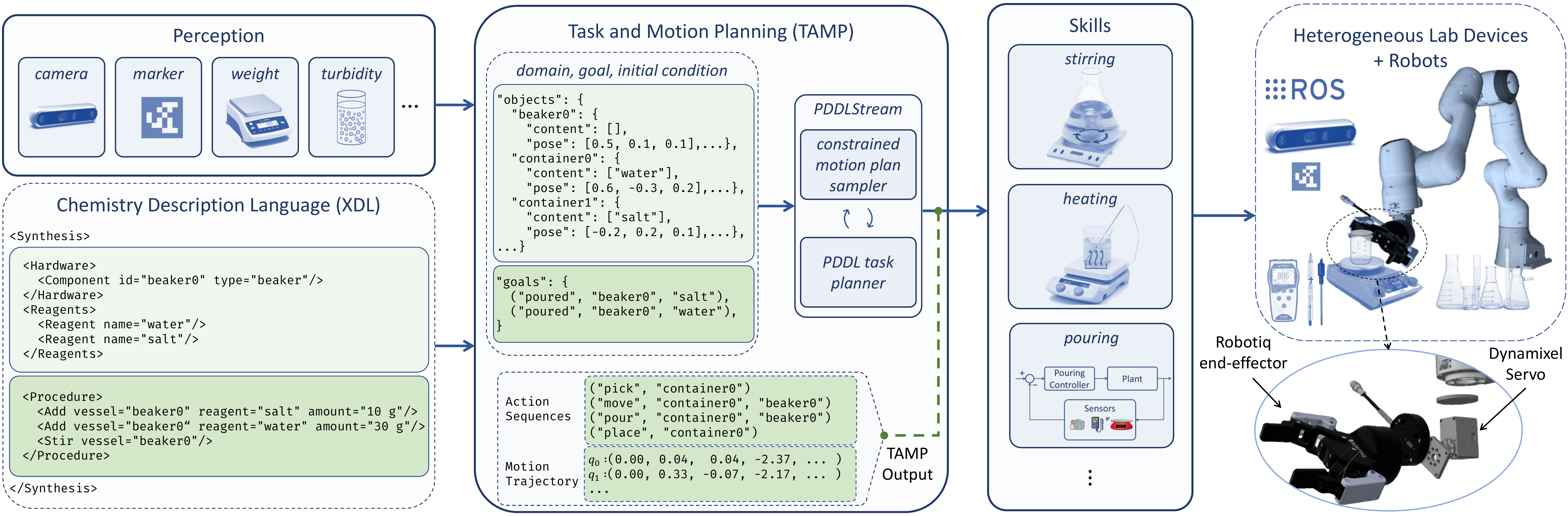}
    \centering
    \caption{\textbf{Framework for robot-assisted chemical synthesis}: composed of \textit{Perception}, \textit{Task \& Motion Planning}, and \textit{Skills} blocks.
    Our framework enables the robot to leverage available chemistry lab devices (including sensors and actuators) by adding them to the robot network through ROS. The robot is equipped with an additional DOF at the end-effector, allowing it to perform constrained motions.
    %  tasks such as ... \AniG{complex scenarios with a high success rate -- vague}.
     Our framework receives the chemical synthesis goal in XDL format. The \textit{procedure} component is converted into corresponding PDDL goals, and 
     \textit{hardware} and \textit{reagents} components identify the required initial condition for synthesis. \textit{Perception} detects objects and estimates their positions, contents in the workspace, and task progress.
     PDDLStream generates a sequence of actions for the robot execution.
    %  The figure uses a solubility experiment (in green) to illustrate usage. 
    %  The framework receives the chemical synthesis goal from a chemist in XDL format, perceives the environment, and eventually commands the robot and lab devices to experiment. An example of a solubility experiment is provided in green color. 
    %  The actions are sent to the robot and executed.
    %  perceives the environment, and eventually commands the robot and lab devices to experiment. 
    % Interpretation of XDL. The XDL Hardware and Reagents components are parsed as objects, and the position of the object obtained from AprilTag is stored in the knowledge base. The XDL Procedure component is converted into corresponding PDDL goals. Using the object positions and goals stored in the knowledge base, PDDLStream generates a sequence of robot actions. The actions are sent to the robot and executed.
    % \AniG{rearrange for this to be two column wide figure if possible. this is the key contribution }
    \vspace{-3pt}
    }
    \label{fig:architecture}
\end{figure*}

% \subsection{Framework Overview}
\paragraph{\textbf{Framework Overview}}
Our proposed framework consists of three components: \textit{perception}, \textit{task and motion planning} (TAMP), and a set of manipulation \textit{skills}, as shown in Fig.~\ref{fig:architecture}.
The Chemical Description Language (XDL) \cite{mehr2020universal} provides a high-level description of experiment instructions as an input to the TAMP solver. 
%Perception modules 
% The TAMP is initialized by parsing a given XDL recipe that contains information about the experiment steps (i.e., the intermediate goals to perform the synthesis), equipment such as vessels, and the required reagents such as water. 
The perception module updates the scene description by detecting the objects and estimating their positions using fiducial markers~\cite{olson2011apriltag}.
Currently, we assume prior knowledge of vessel contents and sizes, and each vessel is mapped to a unique marker ID.
% The detected objects in the scene are  using AprilTag, with
% Based on the AprilTag detection, we can estimate the pose and size of the object.
% This perceptual information are used to update the scene description.
Given the instructions from XDL and the instantiated workspace state information from perception, a sequence of high-level actions and a  robot trajectory are simultaneously generated by the PDDLStream TAMP solver. The resulting plan is then realized by the manipulation module and robot controller, while closing the loop with perception feedback, such as updated object positions and status of the solution.

%\subsection{Task and Motion Planning from XDL}
\subsection{Task and Motion Planning for Chemistry Experiments}
% The robot needs to perform different actions in different scenarios to function in a dynamic environment, in which object positions are not fixed.
% For example, obstacles must be removed before picking up an object if a pick up motion will cause collisions.

% We achieve this through introducing TAMP to increase the adaptability of the robot system.
% The TAMP module generates a sequence of actions and the trajectory of the robot simultaneously.
The TAMP module converts experiment instructions given by XDL into PDDLStream goals and generates a motion plan. The TAMP algorithm is shown in Alg.~\ref{alg:RobotChemsit}. 

\paragraph{\textbf{PDDLStream}}
A PDDLStream problem described by a tuple $(\mathcal{P}, \mathcal{A}, \mathcal{S},\mathcal{O}, \mathcal{I}, \mathcal{G})$ is defined by a set of predicates $\mathcal{P}$, actions $\mathcal{A}$, streams $\mathcal{S}$, initial objects $\mathcal{O}$, an initial state $\mathcal{I}$, and a goal state $\mathcal{G}$.
A predicate is a boolean function that describes the logical relationship of objects.
A logical action $a \in \mathcal{A}$ has a set of preconditions and effects.
The action $a$ can be executed when all the preconditions are satisfied. After execution, the current state changes according to the effects.  The set of streams, $\mathcal{S}$, distinguishes a PDDLStream problem from traditional PDDL. Streams are conditional samplers that yield objects that satisfy specific
constraints. The goal of PDDLStream planning is to find a sequence of logical actions and a continuous motion trajectory starting from the initial state until all goals are satisfied, ensuring that the returned plan is valid and executable by the robot.
We define four types of actions in our PDDLStream domain, including \textit{pick}, \textit{move}, \textit{place}, and \textit{pour}. For example, the \textit{move} action translates the robot end-effector from a grasping pose to a placing or pouring pose using constrained motion planning.
PDDLStream handles continuous motion using streams. Streams generate objects from continuous variables that satisfy specified conditions, such as feasible grasping pose and collision-free motion.
An instance of a stream has a set of certified predicates that expands $\mathcal{I}$ and functions as preconditions for other actions.

A PDDLStream problem is solved by invoking a classical PDDL planner~\cite{helmert2006fast} with optimistic instantiation of streams (Alg.~\ref{alg:RobotChemsit}, line~7).
% $\mathcal{I}$ is expanded by a set of optimistic stream instances $I_{\mathrm{opt}}$.
% The instances in $I_{\mathrm{opt}}$ are optimistic in the sense that their feasibility has not been evaluated yet.
% A PDDL problem $(\mathcal{A}, \mathcal{I} \cup I_{\mathrm{opt}}, \mathcal{G})$ is solved by an arbitrary PDDL solver such as FastDownward~\cite{helmert2006fast}.
If a plan for the PDDL problem is found, the optimistic stream instances $s \in \mathcal{S}$ in the plan are evaluated to determine the actual feasibility (Alg.~\ref{alg:RobotChemsit}, line~8).
%If no plan was found or the streams are not feasible, $I_{\mathrm{opt}}$ is further expanded, and other plans are explored.
If no plan was found or the streams are not feasible, other plans are explored with a larger set of optimistic stream instances.
% We used the adaptive algorithm for optimistic evaluation of a stream.

% We define four types of actions in our PDDLStream domain: pick, move, place, and pour.
% The pick action moves the robot hand to the grasping pose.
% Grasping poses are generated based on the detected object position and the known size of the robot hand.
% The move action moves the robot hand from the grasping pose to the placing pose or pouring pose using constrained motion planning.
% To ensure the robot trajectory is collision-free, we prepare a virtual environment that simulates the collision objects before calling a motion planner.
% The place action opens the gripper to put the grasping object on the table or other objects.
% Lastly, the pour action invokes the pouring skills described in \ref{method-skills}.

\paragraph{\textbf{Chemical Description Language (XDL)}}
XDL~\cite{mehr2020universal} is a chemical description language that describes chemical experiments in a standard XML format.
It is based on XML syntax and is mainly composed of three mandatory sections: \verb|Hardware|,
% the definition of vessels to conduct reaction, 
\verb|Reagents|,
% the definition of reagents used in the experiment
 and \verb|Procedure|.
We parse XDL instructions and pass them to the TAMP module. The \verb|Hardware| and \verb|Reagents| sections are parsed as initial objects $\mathcal{O}$.
\verb|Procedure| is translated into a set of goals $Goals$ (line~1).
$\mathcal{I}$ is generated from $\mathcal{O}$ and sensory inputs (line~2).
Each intermediate goal $\mathcal{G} \in Goals$ is processed by PDDLStream (line~5).
If a plan to attain $\mathcal{G}$ is found, it is stored (line~10) and $\mathcal{I}$ is updated according to the plan (line~11).
After a set of plans to attain all goals is found, we obtain a complete motion plan (line~12). 

\paragraph{\textbf{Task and Motion Plan Refinement at Execution Time}}
We adopt two considerations for the dynamic nature of chemistry experiments: motion plan refinement and task plan refinement.

The generated motion plan is refined to reflect the updated status of the scene and to overcome the perception errors.
The initial object pose detection may contain errors, therefore, the object may not be placed in the expected position during execution. 
This error arises from two reasons. First, when the robot interacts with the objects in the workspace, their position changes, for example when regrasping an object after placing it in the workspace. This change is not foreseeable by the planner ahead of time.
%Second, as proved in App.~\ref{app:perception-error} in the supplementary materials on the website, the perception error is lower when the grasping pose is estimated when the robot in-hand camera is closer to the target object, considering the hand-eye calibration error. Lowering the perception error makes the robot more robust to grasping failures. 
Therefore, to improve the success rate, the object pose is estimated just before grasping, and the trajectory is refined. We assume that the perturbation of the perceived state of the objects is bounded so that it does not cause a change in the logical state of the system, which would necessitate task-level replanning.    

%In addition, to support conditional statements in chemistry experiments, those conditions are met while executing actions. For example, add acid until pH reaches 7. The task execution is repeated using the feedback from perception modules at execution time until conditions are met.

In addition to motion refinement, we consider task plan refinement. Task execution is repeated using the feedback from perception modules at execution time to support conditional operations in chemistry experiments, such as adding acid until pH reaches 7. The number of repetitions required to satisfy conditions is unknown at planning time, so the task plan is refined at execution time.
%includes conditional actions, reconsidering PDDL actions the number of task execution is refined at execution time. to support the conditional statement in chemistry experiments.
%When the experiment specification includes a conditional statement, e.g., , the robot should repeat actions until certain conditions are satisfied.

%As the number of repetitions cannot be specified at the planning time, checking the loop condition can be considered as an execution time refinement. However, I am not sure whether we should describe it here because we just repeat the planned motions.

% \textcolor{blue}{
% Robots have been used to execute experiments specified by human scientists. 
% Mehr \textit{et al}. proposed an automated workflow that translate organic chemistry literature into a structured language called XDL, which can be executed in their hardware called Chemputer~\cite{mehr2020universal}.
% As XDL is high-level description of experiment that is independent of the hardware, we adopt it as the input to our system.
% % We use a task and motion planner to generate motion plans and task plans from high-level abstractions, namely XDL procedures.
%}

\begin{algorithm}[t]
\scriptsize
\begin{algorithmic}[1]
\caption{\textsc{TampForLabAutomation()}}
\label{alg:RobotChemsit}
\Require {A XDL recipe $\mathcal{\chi}$, sensory input $\mathcal{H}$, PDDLStream domain $\mathcal{D}$}
\Ensure {Reference $plan$ to execute}
\State $Goals, \mathcal{O} \leftarrow$\textsc{xdlParser($\mathcal{\chi}$)} \Comment{Objects}
\State $\mathcal{I}  \leftarrow$ \textsc{perception($\mathcal{H}, \mathcal{O}$)}  \Comment{Initial conditions}
\IfThen{not \textsc{passConditions($\mathcal{I}$, $\mathcal{O}$)}} 
    {\Return} 

\State $plan = \emptyset$, $\mathcal{P}=\emptyset $
\ForAll{$\mathcal{G} \in Goals$} 
\While{$time() \leq t_{max} $}
\State $\mathcal{P}$ = \textsc{optimisticPddlStreamPlan($\mathcal{I}$, $\mathcal{G}$, $\mathcal{D}$)}

\IfThen{$\mathcal{P} \neq \emptyset$ and \textsc{isStreamFeasible}($\mathcal{P}$)}{\textbf{break}}
\EndWhile
\IfThen{$\mathcal{P} = \emptyset $}{\Return}
\State $plan \leftarrow plan \cup \mathcal{P}$

\State $\mathcal{I}$ = \textsc{updateSceneRepresentation}($\mathcal{I}, \mathcal{P}$)
\EndFor
\State \Return $plan$

\end{algorithmic}
\end{algorithm}

\subsection{Motion Constraints for Spillage Prevention}% for Chemistry Experiment}
Unlike pick-and-place of solid objects, robots in a chemistry lab need to carry beakers that contain liquids, powders, or granular materials. These chemicals are sometimes harmful, so the robot motion planner should incorporate constraints to prevent spillage. To this end, an important requirement for robot motion is the orientation constraints of the end-effector. To avoid spillage, the end-effector orientation should be kept in a limited range while beakers are grasped. We incorporated constrained motion planning in the framework to meet these safety requirements, under the assumption of velocity and acceleration upper bounds. Moreover, we introduced an additional (8th) degree of freedom to the robot arm, in order to increase the success rate of constrained motion planning. We empirically observed no spillage as long as orientation constraints are satisfied in the regular acceleration and velocity of the robot end-effector, particularly since beakers are not filled to their full capacity in a chemistry lab.

% To avoid the spillage of the beaker contents in chemical experiments during transfer, the robot end-effector velocity and acceleration are limited, and the beaker contents are below the enumerated range printed on it. 

%To ensure that chemicals are not spilled during the manipulation of vessels, constrained motion planning is used to impose hard constraints on the end-effector orientation. 

%\paragraph{\textbf{Planning in a Constrained State Space}}

{
\noindent
\begin{minipage}{0.5\columnwidth}
\begin{algorithm}[H]
\scriptsize
\begin{algorithmic}[1]
\caption{\textsc{ConstrainedMotionPlanning()}}
\label{alg:ConstrainedPathPlanning}
% \Require {$\bm{q}$}
% \Ensure {$\bm{q}_{proj}$}
\ForAll{ $i \in trials$}
\State $\bm{q}_g \leftarrow$ \textsc{solveIK($(^{\mathcal{I}}\bm{p}_{\mathcal{B}}, ^{\mathcal{I}}{\bm{R}}_{\mathcal{B}})$)}
\State \textsc{pathPlanner}$\leftarrow$ init($\bm{q}_0, \bm{q}_g$)
\While{ $path~\text{is}~\emptyset$}
\State $\bm{q} \leftarrow$ \textsc{sample()}
\boxitt{light_red}{3.2}
\While{$\| \bm{\mathcal{F}}(\bm{q})\| > \epsilon$} %\Comment{Projection}

\State $\delta \bm{q} \leftarrow \bm{\mathcal{J}}^\dagger(\bm{q}) \bm{\mathcal{F}}(\bm{q})$
\State $\bm{q} \leftarrow \bm{q} - \delta \bm{q}$
\EndWhile
\State $path \leftarrow$ \textsc{pathPlanner($\bm{q})$}
\EndWhile
\IfThen{$path \neq \emptyset $}{\Return} $path$
\EndFor
\State \Return $path$
\end{algorithmic}
\end{algorithm}
\end{minipage}%
\hfill
\begin{minipage}{0.4\columnwidth}
\vspace{1em}
{\small \textbf{Constrained motion planning:} First, an IK solver is executed with multiple random initializations ~\cite{Beeson_Ames_2015}. Then, PRM$^\star$ is initialized \cite{karaman2011sampling}. PRM$^\star$ graph grows using projected samples that satisfy the constraints (red). This process iterates until a motion plan is found.
}
\end{minipage}
}

\vspace{0.5em}

\paragraph{\textbf{Constrained Motion Planning}}
% Traditionally, the unconstrained planning problem can be described by finding a path in the manipulator's free configuration space ($\mathcal{Q}_{free} \in \mathbb{R}^n$) that satisfies a given start state $q_{start}$ and goal state $q_{goal}$ without entering the obstacle subspace $\mathcal{Q}_{obs}$.
% initial configuration $\bm{q}_0 \in \mathbb{R}^n$
% , end-effector goal pose $(^{\mathcal{I}}\bm{p}_{\mathcal{B}}\in \mathbb{R}^3, ^{\mathcal{I}}{\bm{R}}_{\mathcal{B}}\in SO(3))$ ,
% 
Given a robot   with $n$ degrees of freedom in the workspace $\mathcal{Q} \in \mathbb{R}^n$ with obstacle regions $\mathcal{Q}_{obs} \in \mathbb{R}^n$, the constrained planning problem can be described by finding a path in the manipulator's free configuration space $\mathcal{Q}_{free}= \mathcal{Q} -\mathcal{Q}_{obs} $ that satisfies initial configuration $\bm{q}_0 \in \mathbb{R}^n$, end-effector goal pose $(^{\mathcal{I}}\bm{p}_{\mathcal{B}}\in \mathbb{R}^3, ^{\mathcal{I}}{\bm{R}}_{\mathcal{B}}\in SO(3))$, and equality path constraints $\bm{\mathcal{F}}(\bm{q}): \mathcal{Q} \to \mathbb{R}^k$.
The constrained configuration space can be represented by the implicit manifold $\mathcal{M} = \{ q \in \mathcal{Q} \ | \ \bm{\mathcal{F}}(\bm{q}) = \textbf 0 \}$. 
The implicit nature of the manifold prevents planners from directly sampling, since the distribution of valid states is unknown. 
Further, since the constraint manifold resides in a lower dimension than the configuration space, sampling valid states in the configuration space is highly improbable and thus impractical.
Following the constrained motion planning framework developed in \cite{Kingston2019Exploring, kingston2018sampling}, our framework integrates the projection-based method for finding constraint-satisfying configurations during \textit{sampling} as described in Alg.~\ref{alg:ConstrainedPathPlanning}. 
In this work, the constraints are set to the robot end-effector, hence they can be described with geometric forward kinematics, with its \textit{Jacobian} defined as $\bm{\mathcal{J}}(\bm{q})= \frac{\delta \bm{\mathcal{F}}}{\delta \bm{q}}$. After sampling from $\mathcal{Q}_{free}$ in line 5, projected configurations $\bm{q}$ are found by minimizing $\bm{\mathcal{F}(q)}$ iteratively using Newton's method (highlighted in red). 
% Given a constraint on the geometry of the robot that lowers the dimensionality, such as an end-effector orientation constraint, the planning problem resides on a constraint manifold $\mathcal{M}$ embedded inside the original configuration space $\mathcal{Q}$. 
% For some function of the joint state $\textbf F(q): \mathcal{Q} \to \mathbb{R}^k$ derived from the $k$ specified constraints, the constraint satisfaction condition is defined by the equality $\textbf F(q) = \textbf 0$.
% Then, the constrained configuration space can be represented by the implicit manifold $\mathcal{M} = \{ q \in \mathcal{Q} \ | \ \textbf F(q) = \textbf 0 \}$. 
% The implicit nature of the manifold prevents planners from directly sampling, since the distribution of valid states is unknown. 
% Further, since a manifold resides in a lower dimension than the configuration space, sampling valid states in the configuration space is highly improbable and thus impractical.
% Following the constrained motion planning framework developed in \cite{doi:10.1177/0278364919868530}, our system integrates the projection-based method for finding constraint-satisfying configurations into the TAMP module as part of multiple streams. 
% Using traditional sampling-based motion planning methods, we use a projection operator $P: \mathcal{Q} \to \mathcal{M}$ to map sampled configurations in $\mathcal{Q}$ to valid configurations in $\mathcal{M}$ while preserving asymptotically optimal guarantees.
We use probabilistic roadmap methods (PRM$^\star$) to plan efficiently in the 8-DoF configuration space found in our chemistry laboratory domain \cite{karaman2011sampling, kavraki1996probabilistic}.%PRM$^\star$ is a multi-query planner that constructs a persistent probabilistic roadmap of $\mathcal{Q}_{free}$ in a separate thread, allowing for multiple motion planning queries to be solved without re-exploring the environment.
% In practice, the PDDLstream actions of \verb|grasp| and \verb|pre-grasp| require less than 5 seconds to find a plan, since a large roadmap has already been constructed from the previously requested \verb|pour| action that moves across the workspace.
% \paragraph{\textbf{Inverse Kinematics for Constrained Motion Planning}}

The constrained path planning problem is sensitive to the start and end states of the requested path, since paths between joint states may not be possible under strict or multiple constraints. 
% Since the task plan only specifies the positions of objects in Cartesian space, the start and goal states of the motion plan must be derived through inverse kinematics (IK).
If constrained planning is executed with any arbitrary valid solution from the IK solver, the planner typically fails. 
To address this shortcoming, three considerations are made. First, a multi-threaded IK solver with both iterative and random-based techniques is executed, and the solution that minimizes an objective function $\phi$ is returned \cite{Beeson_Ames_2015}. During grasping and placing, precision is paramount, and we only seek to minimize the sum-of-squares error between the start and goal Cartesian poses. 
Second, depending on the robot task, the objective function is extended to maximize the manipulability ellipsoid 
%, $\phi = \max \sqrt{\det(JJ^T)}$,
as described in \cite{doi:10.1177/027836498500400201}, which is applied for more complicated maneuvers, such as transferring liquids across the workspace. 
% Finally, note that configuration sampling must account for the fact that multiple goal poses are possible; for example, any end-effector yaw angle is valid during parallel grasping but roll and pitch angles should be zero.
Finally, note that configuration sampling must account for the fact that multiple goal configurations are possible. For this purpose, Alg.~\ref{alg:ConstrainedPathPlanning} can iterate several times to find various goal configurations in line 2.

\paragraph{\textbf{An 8-DoF Robot Arm}}
To increase the success rate of planning and grasping under non-spillage constraints, we introduced an additional degree of freedom to the 7 DoF Franka robot.
The aim of the addition is twofold.
First, the 8-DoF robot has a higher empirical success rate in constrained motion planning, which leads to a higher success rate in total task and motion planning.
% \textcolor{blue}{\textbf{This is proven in App.\ref{}.}}
Second, the robot end-effector orientation is changed to flat (parallel to the floor). This usually places the end-effector orientation far from the joint limit, which in turn makes the pouring control easier. % We should have the data to support this argument.

\subsection{Manipulation and Perception Skills}
\label{method-skills}
Chemistry lab skills require a particular suite of sensors, algorithms, and hardware. We provide an interface for instantiating different skill instances through ROS and simultaneously commanding them. For instance, recrystallization experiments in chemistry require both pouring, heating, and stirring, which uses both weight feedback for volume estimation and skills for interacting with the liquid using available hardware.

\paragraph{\textbf{Pouring Controller}}
In chemistry labs, a frequently used skill in chemical experiments is pouring. %\textcolor{blue}{for synthesizing new materials} is pouring.
Pouring involves high intra-class variations depending on the underlying objective (e.g. reaching a desired weight or pH value); the substances and material types being handled (e.g. granular solids or liquids); the glassware being used  (e.g., beakers and vials); the overall required precision; and the availability of accurate and fast feedback. Pouring is a closed-loop process, in which feedback should be continuously monitored.
% The type of sensor used for measurement depends on the pouring objective. For example, in the case of adding a specific amount of content to a target vessel, then weight or visual feedback can be used to estimate the content volume. 
% In another example of a titration process, rapid color or pH changes should be measured.
Among these pouring actions, in our work we consider the following variations: pouring of liquids and granular solids.
% powders, and operating squeeze bottles. %and opening/closing burette valves. 
% In all these skills, chemists use their visual information as well as sensory readouts, for example, scale weight readouts, to perform the pouring action. Those sensory readouts are directly related to pouring objective. For example, in case of pouring liquid, the objective is to fill a container with a specified amount of liquid (in mL or in grams). In another example of titration, the pouring goal is directly related to the pH value of a solution and observed by rapid color changes.
Note that, in contrast to many control problems, pouring is a non-reversible process as we cannot compensate for overshoot (as the poured material cannot go back to the pouring beaker). 

\begin{figure}[t]
    \includegraphics[width=0.8\columnwidth]{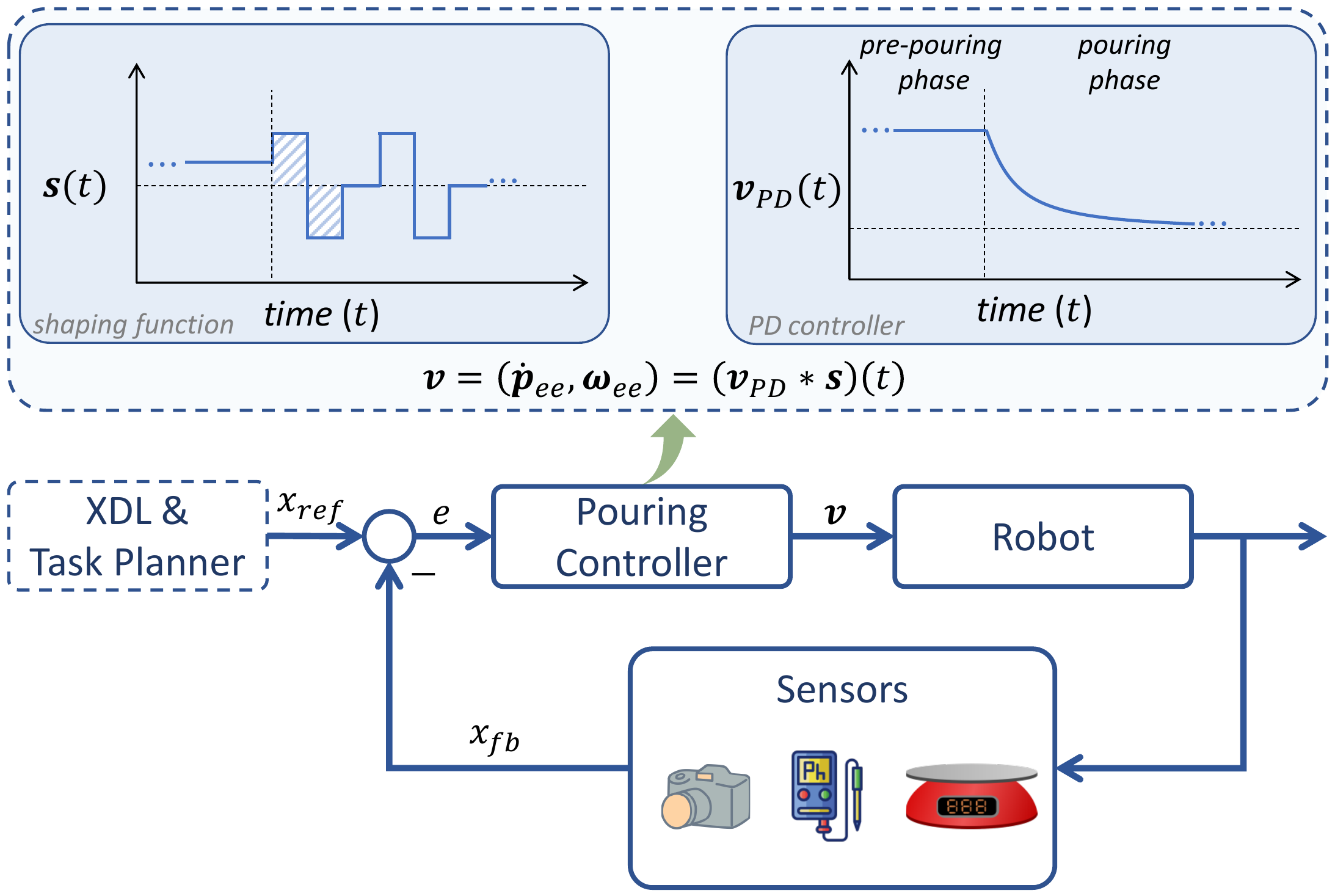}
    \centering
    \caption{\textbf{Pouring skill controller}: given the XDL \& TAMP reference values and sensor feedback, the pouring controller computes the end-effector velocity for the robot using the convolution of a shaping function $\bm{s}(t)$ and a PD control output $\bm{v}_{PD}(t)$.}
    \label{fig:skills}
\end{figure}

\begin{figure}[t]
\begin{minipage}{0.65\linewidth}
    \centering
    \includegraphics[width=0.9\linewidth]{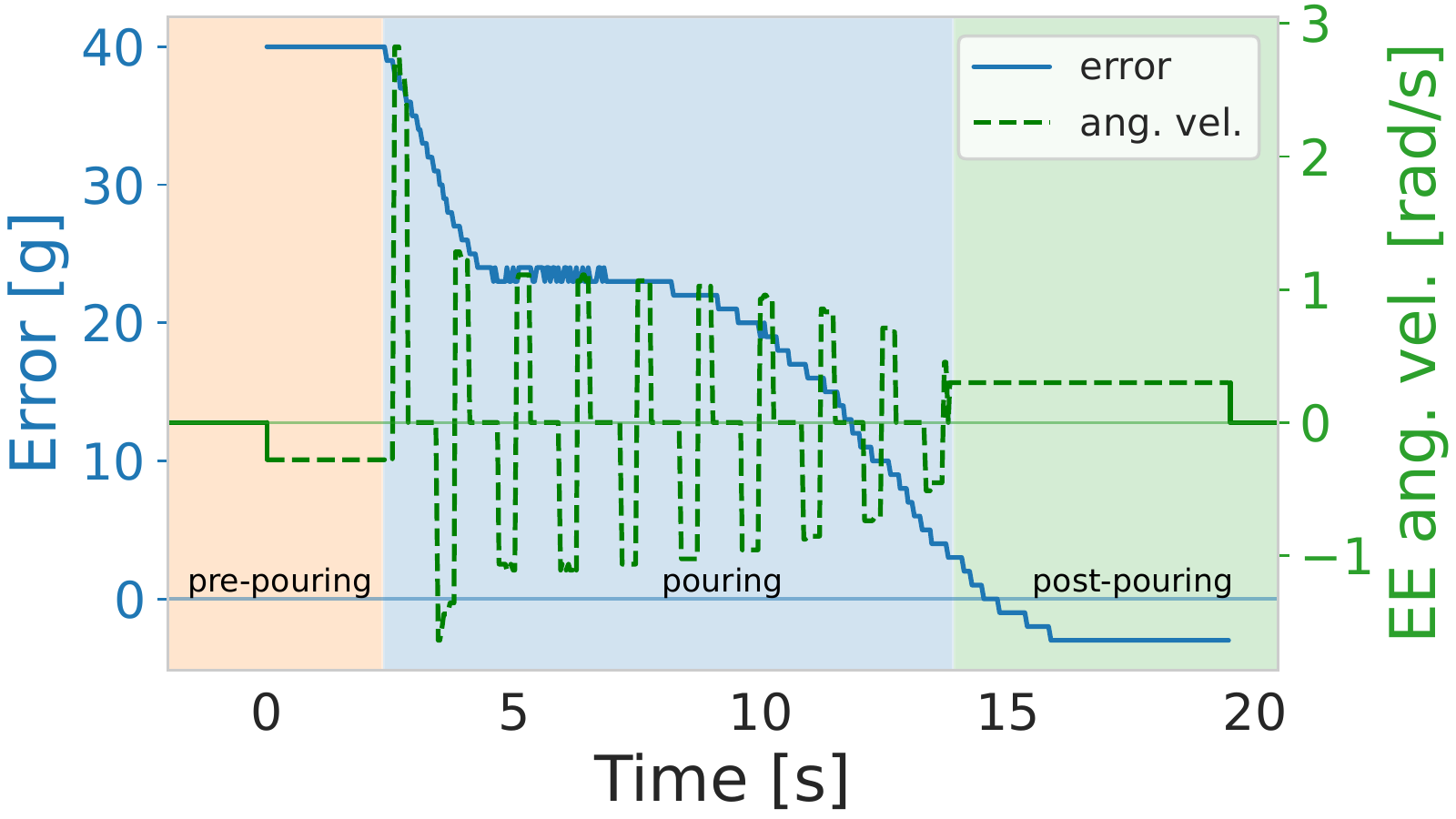}
    \end{minipage}
    % \vspace{-1cm}
\begin{minipage}{0.30\linewidth}
    \centering
    \caption{\textbf{An example of pouring control.} The velocity of the end-effector is controlled based on the feedback error and shaping function.
    % \vspace{-1cm}
    }
    \label{fig:pour_control}
\end{minipage}
\end{figure}

Inspired by observations of chemists pouring reagents, we propose a controller that allows the robot to perform different pouring actions.
% The proposed approach is based on classical control techniques.
As shown in Fig.~\ref{fig:skills}, the proposed method takes sensor measurements (e.g. weight feedback from the scale) as feedback and a reference pouring target. The algorithm outputs the robot end-effector joint velocity describing oscillations of the arm's wrist. 
% Pouring is composed of two phases: the \textit{pre-pouring phase}, when the end-effector starts moving but the measurements are fixed (i.e., the vessel rotates until its content reaches the beaker edge), and the \textit{pouring phase}, when the robot stays in the proximity of the pouring threshold to overflow in a controlled fashion (sensor measurement changes).
Since sensors are characterized by measurement delays, chemical reactions require time to stabilize, and pouring is a non-reversible action, chemists tend to conservatively pour a small amount of content from the pouring vessel into the target vessel. They periodically wait for some time to observe any effects and then pour micro-amounts again. In our approach, we use a shaping function $s(t)$ to guide the direction and frequency of this oscillatory pouring behavior, while a PD controller lowers the pouring error.
% , guiding the general behavior of the end-effector.
The end-effector velocity vector is computed by convolving the shaping function $s(t)$ over the PD control signal, $\bm{v}_{\mathrm{PD}}(t)= \bm{k}_p e + \bm{k}_d \dot{e}$, where $e(t)= x_{ref} - x_{fb}$. 
Fig.~\ref{fig:pour_control} shows an example of the angular velocity of the end-effector and the error during actual pouring.
% More information about and videos of the pouring method can be found on the project website for this work.
% This process has been replicated by the shaping function $s(t)$, where its parameters can be tuned according to the desired pouring accuracy, efficiency, content, vessel type, vessel shape, and feedback sensors. These parameters are currently tuned manually. In the future, we will tune them automatically using simulation tools, learning techniques, and sensor fusion. 

% PD controller with scale feedback, joint velocity control (feedback control), parallel grasping

\begin{figure}[t]
\begin{minipage}{0.37\linewidth}
    \centering
    \includegraphics[width=0.9\linewidth, trim={5mm 10mm 5mm 10mm}, clip]{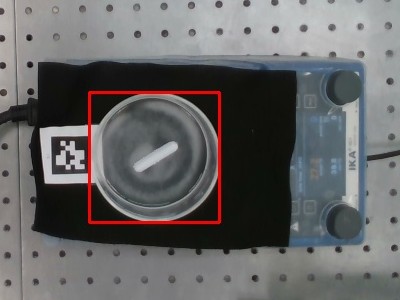}
    \end{minipage}
\begin{minipage}{0.56\linewidth}
    \centering
    \caption{\textbf{An example of automated turbidity measurement.} The camera detects the Petri dish using Hough Circle Transform. The average brightness of the detected area (red square) is used as a proxy of turbidity.}
    \label{fig:dish_detection}
\end{minipage}
\end{figure}

% \begin{figure}
% \begin{minipage}{0.37\linewidth}
%     \centering
%     \includegraphics[width=\linewidth]{images/v_channel.pdf}
%     \end{minipage}
% \begin{minipage}{0.56\linewidth}
%     \centering
%     \caption{\textbf{Turbidity calculation.} V channel of the region of the dish is shown (a) before dissolving, (b) after dissolving.}
%     \label{fig:v_channel}
% \end{minipage}
% \end{figure}

\paragraph{\textbf{Turbidity-based Solubility Measurement}}
Solubility of a solute is measured by determining the minimum amount of solvent (water) required to dissolve all solutes at a given temperature when the overall system is in equilibrium \cite{shiri2021automated}.
%\st{We pour a specified amount of solute into the Petri dish first, and repeatedly pouring 10 mL of water into the dish until all solutes get dissolved.}
Since the solutions get transparent when all solutes dissolve into water, turbidity, or opaqueness of the solution, is used as the metric to determine the completion of the experiment.
The average brightness of the solution was used as a proxy for the relative turbidity, inspired by HeinSight~\cite{shiri2021automated}.
It compared the current measured turbidity value with reference value (coming from pure solvent) to determine when the solution is dissolved. Differently from them, we use the relative turbidity changes using the current and previous measurement values to detect when the solution is dissolved.
% the equilibrium feature of solubility, i.e., when the solution is fully dissolved the turbidity also reaches an equilibrium value. 
Moreover, to make the perception pipeline  autonomous, when the robot with an in-hand camera observes the dish containing the solution, it detects the largest circular shape as the dish using Hough Circle Transform implemented in OpenCV.
The square region containing the dish is converted into HSV color space, and the average Value (brightness) of the region is used as a turbidity value.
% When the robot observes the solution, the robot moves the in-hand camera above the dish containing the solution and detects the largest circular shape as the dish using Hough Circle Transform implemented in OpenCV.
% The square region containing the dish is converted into HSV color space, and the average Value (brightness) of the region is used as a turbidity value.
Figure~\ref{fig:dish_detection} shows an example of the automated turbidity measurement.
Although the detected area contains the dish and stir bar, they do not affect the relative value because these are a constant bias in all measurements.

% \textcolor{blue}{When all solutes have been dissolved in water, the turbidity does not change even if more water is added.
% We measure the turbidity of the solution right after pouring water and compare the turbidity value after stirring.
% The turbidity during stirring is ignored because the turbidity value increases because of the water current and bubbles during stirring.
% If the turbidity difference is smaller than an arbitrary low threshold percentage, we determine all solutes have been dissolved and we stop the experiment.}
\section{Experiments and Evaluation}
\label{sec:experiments-evaluation}
We evaluate the proposed framework with two component studies on pouring and constrained TAMP, and two types of experiments, solubility and recrystallization.
 
% solubility test and recrystallization, two common multi-step processes in 

\subsection{Experiment Setup}
\paragraph{\textbf{Hardware}}
The proposed lab automation framework has been evaluated using the Franka Emika Panda arm robot, equipped with a Robotiq 2F-85 gripper and an Intel RealSense D435i stereo camera mounted on the gripper to allow for active vision. The robot's DoF has been extended by one degree (in total 8 DoF) at its end-effector using a Dynamixel XM540-W150 servo motor. Fig.~\ref{fig:architecture} shows the hardware setup.

\paragraph{\textbf{Lab Tools Integration}}
The robot framework is expanded by incorporating lab tools. % designed for chemical experiments.
We used an IKA RET control-visc device, which works as a scale, hotplate, and stir plate, and Sartorius BCA2202-1S Entris, which works as a high-precision weighing scale.
The devices communicate with the TAMP solver to execute chemistry specific skills.
%The robot can incorporate available lab tools to extend its abilities for performing chemical experiments, as indicated in Fig.~\ref{fig:architecture}. To perform the solubility and recrystallization experiments, we use an IKA RET control-visc device, commonly found in chemistry labs, to extend the robot skill set. This device works as a scale, hotplate, and stirplate. The scale precision is 1g with approximately $\approx3$ s of measurement delay.
%The device functionalities are exposed through the common skill interface and commutation from the robot to the device is instantiated via a RS-232 Serial to USB adapter.

\paragraph{\textbf{Software}}
The robot is controlled using FrankaPy~\cite{zhang2020modular}. We implemented a ROS wrapper for the servo motor (8th DoF). To detect fiducial markers, we use the AprilTag library~\cite{olson2011apriltag}.
We use the MoveIt motion planning framework~\cite{coleman2014reducing} for our TAMP solver and its streams.
% We use the Open Motion Planning Library (OMPL)~\cite{sucan2012the-open-motion-planning-library}, the MoveIt motion planning framework~\cite{coleman2014reducing}, and TRAC-IK~\cite{Beeson_Ames_2015} for our TAMP solver and its streams.
The constrained planning function~\cite{Kingston2019Exploring} is an extension of elion~\cite{elion}.

\paragraph{\textbf{Chemistry Experiments}} We evaluated our framework by conducting solubility measurement experiments as well as recrystallization experiments.
Solubility is a basic property of the solute and solvent, and measuring it is a well-known basic experiment in chemistry~\cite{wolthuis1960determination}.
Measuring solubility has desirable characteristics as a benchmark for automated chemistry experiments: (i) it requires basic chemistry operations, such as pouring, solid dispensing, and observation of the solution status, (ii) solubility can be measured using ubiquitous food-safe materials, such as water, salt, sugar, and (iii) the accuracy of the measurement can be evaluated quantitatively by comparing with literature values.
We also conducted a recrystallization experiment to show the flexibility of our framework and its ability to handle a diverse set of experiments.
We attained $7.2\%$ error for salt solubility and successfully recrystallized alum.

\subsection{Safety Evaluation of Motions Without Spillage}
\paragraph{\textbf{Constrained Motion Planning in 7/8 DoF robot}}
%In this section, a successful \textit{constrained motion plan} refers to a trajectory such that the end-effector is restricted to a set orientation with some tolerance. Additionally, this section will primarily use \textit{position} goals since our end-effector is restricted to some unit quaternion orientation $\mathbf{Q}$ unless specified otherwise. 
The constrained motion planning performance of 7 DoF and 8 DoF robot is %the unaltered Franka Emika Panda arm (7 DoF) and the extended arm (8 DoF) is 
evaluated in two scenarios: (1) single step, (2) two steps.
In scenario (1), robots find a constrained path with a fixed orientation from initial to final positions that are randomly sampled. Scenario (2) extends the first with an additional intermediate sampled waypoint. 
%Scenario \textit{iii} simulates the transfer and pouring actions during experiments, i.e., two positions are randomly sampled, and the robots go from initial to final positions with the fixed orientation, and later, rotate end-effector \textcolor{blue}{90$^\circ$} while keeping the position.
%Given $n$ tasks, where a task is defined as finding a constrained motion plan from position $\mathbf{p}^1_i$ to $\mathbf{p}^2_i$ such that $\mathbf{p}^1_i,\mathbf{p}^2_i\in \mathbb{R}^3$ are randomly sampled positions.
%In this setup, planning success is evaluated between two poses, and therefore the starting and target joint states can be any Inverse Kinematics (IK) solution. 
For each scenario, we run 50 trials in Alg.~\ref{alg:ConstrainedPathPlanning} with random seeding of the IK solver.
%Each trial involves computing a different IK solution for $\mathbf{s}_i$ and $\mathbf{t}_i$, and then passing the resulting joint states to the constrained planner.
%2.) Chemistry experiments often require multiple trajectories in succession. With this in mind, a task is defined as finding a plan for a two-step sequence $\mathbf{p}^1_i\rightarrow\mathbf{p}^2_i\rightarrow\mathbf{p}^3_i$, where $\mathbf{p}^1_i,\mathbf{p}^2_i,\mathbf{p}^3_i\in \mathbb{R}^3$ are randomly sampled positions. 
%In the sequence, the trajectories $\mathbf{p}^1_i\rightarrow\mathbf{p}^2_i$ and $\mathbf{p}^2_i\rightarrow\mathbf{p}^3_i$ are constrained. Note that step one of the sequence ($\mathbf{p}^1_i\rightarrow\mathbf{p}^2_i$) is analogous to Case 1.
%In step two of the sequence ($\mathbf{p}^2_i\rightarrow\mathbf{p}^3_i$), the planner is restricted to finding a trajectory from a specific starting joint state (the IK solution for $\mathbf{p}^2_i$ from step one) to the next position. The joint state of the next position can be any IK solution. When solving the sequence, every step (e.g. $\mathbf{p}^1_i\rightarrow\mathbf{p}^2_i$) is allowed one trial in Alg.\ref{alg:ConstrainedPathPlanning}. If step one succeeds at finding a plan then move to step two with the resulting joint state.
In scenario (2), we restart the sequence planning from the first step if a step fails. Constraints are set to the robot end-effector pitch and roll ($\| \theta, \phi \| \leq 0.1\text{ rad}$).
% We allow 50 failures.
%3.) To evaluate the pouring ability of the robot arms, we define a task as a constrained motion plan leading into a pouring motion. The task sequence can be described as $\mathbf{p}^1_i\rightarrow\mathbf{p}^2_i\rightarrow\mathbf{P}^{pour}_i$ where $\mathbf{p}^1_i,\mathbf{p}^2_i\in \mathbb{R}^3$ are randomly sampled positions and $\mathbf{P}^{pour}_i=[{\mathbf{p}^2}^\mathsf{T}_i, \mathbf{Q'}^\mathsf{T}]^\mathsf{T}$ is a pouring pose such that $\mathbf{Q'}$ is offset by $90^{\circ}$ around the yaw axis from $\mathbf{Q}$.
%\textcolor{blue}{The pouring motion $\mathbf{p}^2_i\rightarrow\mathbf{P}^{pour}_i$ is treated as an unconstrained motion planning problem such that $\forall j$ $\|\mathbf{p}^{traj}_j-\mathbf{p}^2_i\|<\epsilon$ where $\mathbf{p}^{traj}_j$ is the position of waypoint $j$ in the pouring trajectory. The trial setup used for this case is the same as case 2.}

%IK solutions are found using the TRAC-IK library~\cite{Beeson_Ames_2015}. Given a seed value for joints $\mathbf{q}_{seed}$, the TRAC-IK solver starts the iterative process toward the goal by calculating the Jacobian $J$ of the Cartesian error between the seed pose (computed using \textit{Forward Kinematics}) and target pose. Although TRAC-IK also employs \textit{random restarts} to avoid local minima, a different $\mathbf{q}_{seed}$ can result in a different IK solution for the same target pose. 

The performance of the 7-DoF and 8-DoF robot arms for the two scenarios are shown in Table~\ref{tab:constrained-motion-plan}. The results show that the IK and constrained motion planning have higher success rates in 8-DoF compared with the 7-DoF robot.%solver is able to find valid joint states for all sampled positions for the 8 DoF arm.

% \begin{table}[ht]
% \centering
% \caption{\textbf{Comparison of 7 and 8 DoF robot in constrained motion planning.}
% The robotic arms were evaluated using the same set of sampled points. \textit{IK success} is the percentage of pairs or tuples for which IK solutions are found for the entire sequence. \textit{Plan success} is the percentage of pairs or tuples for which a constrained plan was found. 
% }
% \setlength{\tabcolsep}{1pt}
% \label{tab:cmp}
% \begin{tabular}{p{1cm} p{1.5cm} p{2cm} p{2cm}} \toprule
% \textbf{Robot} & \textit{IK success}[\%] & \textit{Plan success} [\%] & \textbf{Other info} \\ \hline
%  \rowcolor[rgb]{0.906,0.902,0.902}    7 DoF & 99 & 84 & \\ \hline
%                                       8 DoF & 100 & 97 & \\ \hline
% \end{tabular}
% \end{table}
\begin{table} [t]
\centering
  \setlength\belowcaptionskip{-1.7\baselineskip}
\caption{\textbf{Success rate comparison of 7 and 8-DoF robot.}
%Success rates in IK and constrained motion planning are shown. %\textit{IK success} is the percentage of pairs or tuples for which IK solutions are found for the entire sequence. \textit{Plan success} is the percentage of pairs or tuples for which a constrained plan was found.
% and roll ($\| \phi \| \leq 0.1\text{ rad}$) axes.
}
% \begin{tabular}{lSSSSSS}
%     \toprule
%     \multirow{2}{*}{} &
%       \multicolumn{2}{c}{\textit{\textbf{Scenario 1 (\%)}}} &
%       \multicolumn{2}{c}{\textit{\textbf{Scenario 2 (\%)}}} &
%       \multicolumn{2}{c}{\textit{\textbf{Scenario 3 (\%)}}} \\
%       \cmidrule(l){2-3} \cmidrule(l){4-5} \cmidrule(l){6-7}
%       & \textit{\textbf{{IK}}} & {\textit{\textbf{Plan}}} & {\textit{\textbf{IK}}} & {\textit{\textbf{Plan}}} & \textit{\textbf{{IK}}} & {\textit{\textbf{Plan}}}\\
%       \hline
%  \rowcolor[rgb]{0.906,0.902,0.902}   7 DoF & 99 & 84 & 80 & 30 & 99 & 69\\ \hline
%     8 DoF & 100 & 97 & 100 & 84 & 100 & 80 \\
%     \bottomrule
%   \end{tabular}
  \begin{tabular}{lcccc}
    \toprule
    \multirow{2}{*}{} &
      \multicolumn{2}{c}{\textit{\textbf{Scenario 1 (\%)}}} &
      \multicolumn{2}{c}{\textit{\textbf{Scenario 2 (\%)}}} \\
      \cmidrule(l){2-3} \cmidrule(l){4-5}
      & \textit{\textbf{{IK}}} & {\textit{\textbf{Plan}}} & \textit{\textbf{{IK}}} & {\textit{\textbf{Plan}}}\\
      \hline
 \rowcolor[rgb]{0.906,0.902,0.902}   7 DoF & 99 & 84 & 99 & 70\\ \hline
    8 DoF & 100 & 97 & 100 & 84 \\
    \bottomrule
  \end{tabular}
\label{tab:constrained-motion-plan}
\end{table}

\subsection{Pouring Skill Evaluation}
\begin{figure}[t]
\centering
\begin{subfigure}{0.47\columnwidth}
    \includegraphics[width=\textwidth]{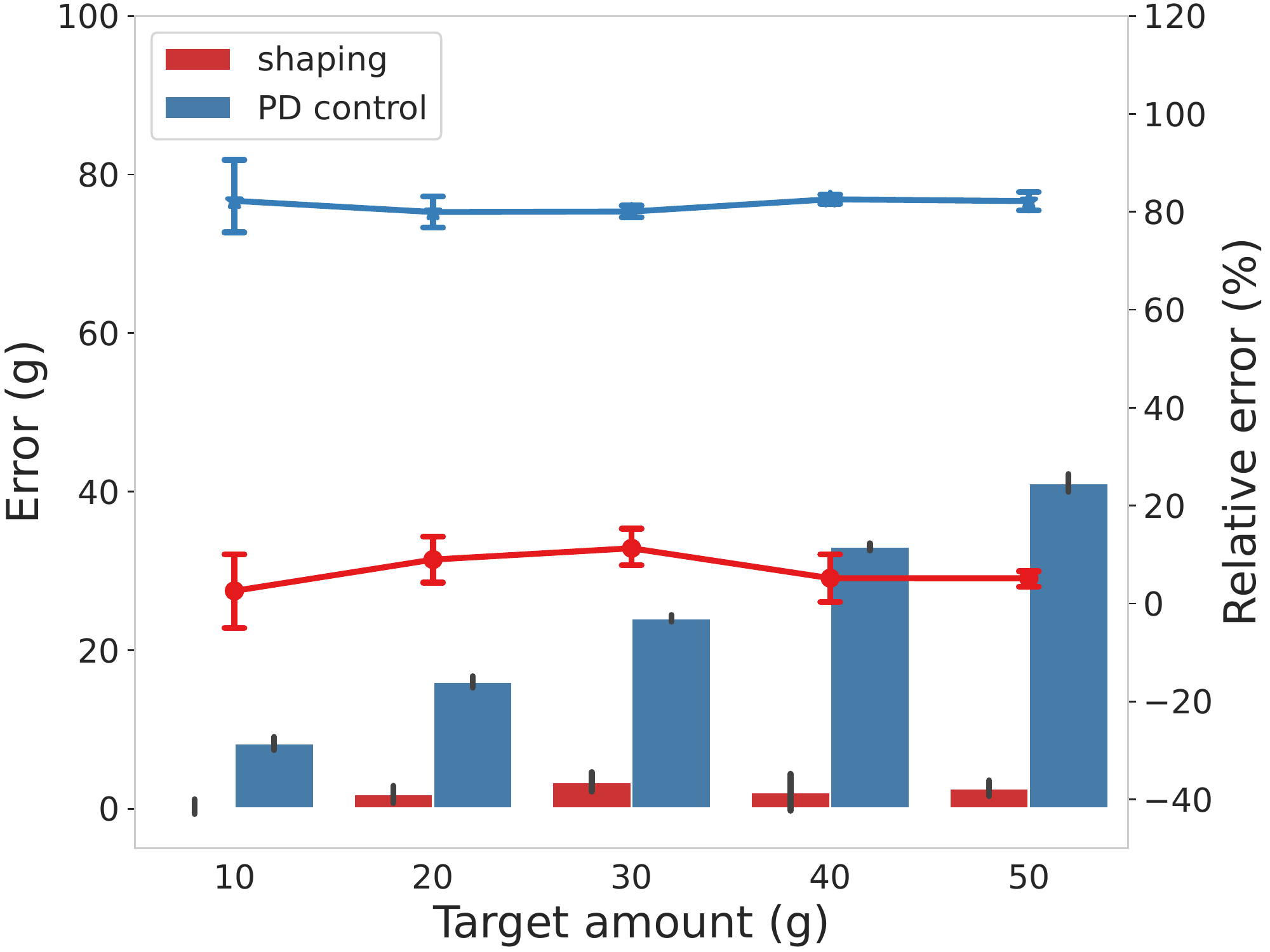}
    \caption{Pouring error in water.}
    \label{fig:pour_error_water}
    \end{subfigure}
    \hspace{0.01\textwidth}
    \begin{subfigure}[b]{0.47\columnwidth}
    \includegraphics[width=\textwidth]{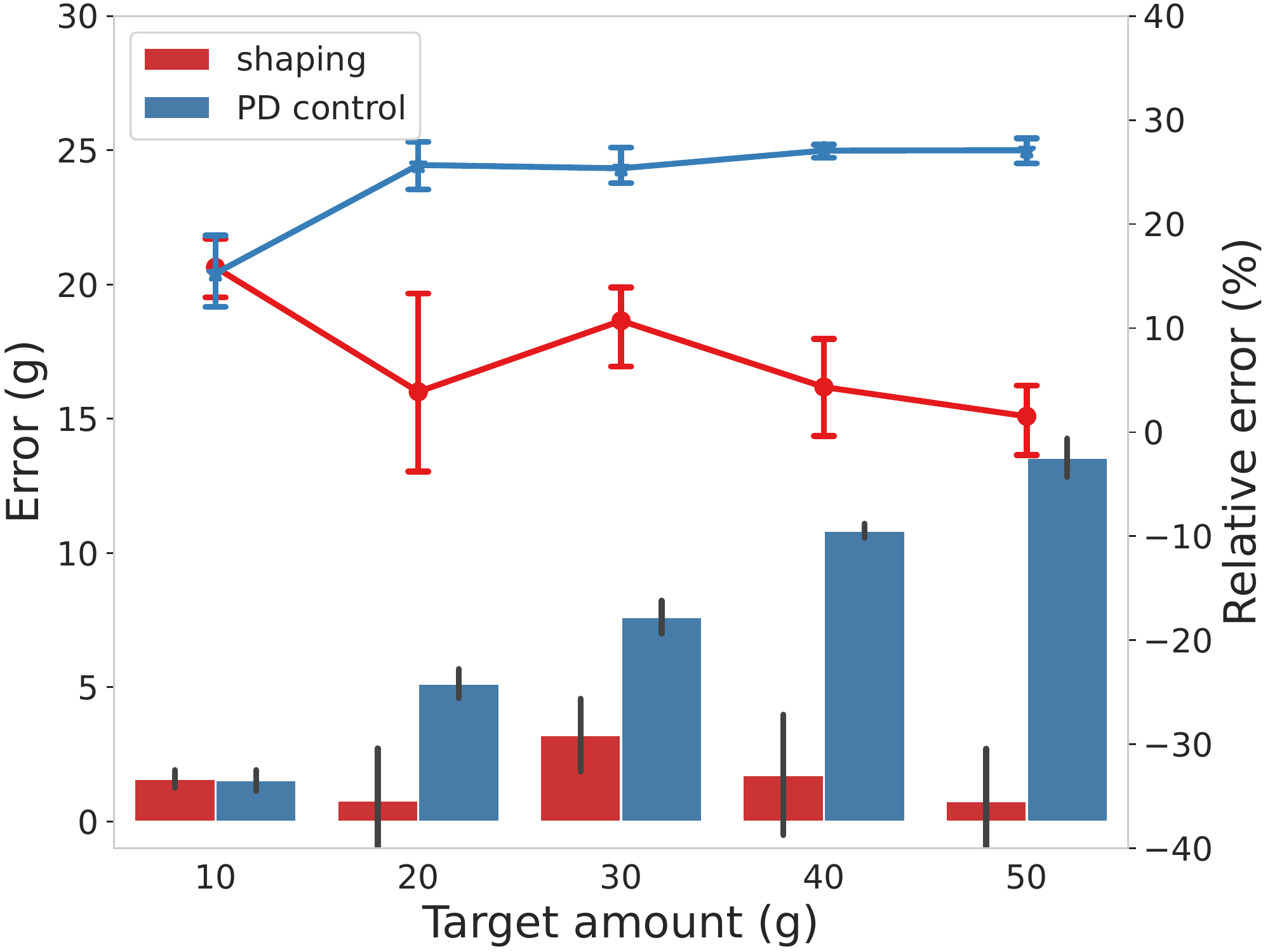}
    \caption{Pouring error in salt.}
    \label{fig:pour_water_salt}
    \end{subfigure}
    \caption{\textbf{Evaluation of pouring error}. The pouring errors of our shaping pouring and PD control baseline pouring are compared using (a) water and (b) salt. The bar plot shows the error (poured amount - target amount) and the line plot shows the relative error. The error bars show the standard deviation. \vspace{-1mm}}
    \label{fig:pouring_error}
\end{figure}

We evaluated accuracy and efficiency of pouring skill of liquid and powder.
To evaluate the effect of our proposed pouring method, we implemented a PD control pouring method where end-effector angular velocity is proportional to the difference between target and feedback weight as a baseline.
Fig.~\ref{fig:pouring_error} shows the pouring experiment results.
The results show that the shaping function contributed to reducing the overshooting compared to PD control pouring.
The overshoot of the PD control pouring is mainly because of the scale's delayed feedback ($\sim$3 s).
The intermittent pouring caused by the shaping function compensated for the delay and improved the overall pouring accuracy.
% \textcolor{blue}{SHOULD WE COMPARE THE RESULTS WITH THE SOA PAPERS, WE ARE CLOSE TO THEM; IF WE USE PRECISE SCALE WE MAY HAVE EVEN A BETTER RESULT; MENTIONED IN NOVELTY; what about efficiency? We may report them; even better than them}
On average the pouring error using shaping approach is $2.2 \pm 1.5$ g and for PD control is $24.5 \pm 12.0$ g and their average relative error and standard deviation are $8.1 \pm 4.8$ \% and $81.4 \pm 4.5$ \%. Moreover, as we can see both the error and relative errors stays constant when using shaping method in contrast to the PD controller.
The average pouring time with shaping function of 50 mL water and salt were 25.1 s and 36.8 s.
Our results are comparable with previous work~\cite{Kennedy2019Autonomous, Huang2021Robot} in terms of pouring error and time, without using a learned, vision-based, policy, or expensive equipment setup.
% The pouring skill evaluation is shown in Fig.~\ref{fig:pouring_error}.
% The accuracy of pouring is measured by the final error.
% The average absolute error is 2$\pm$1 g for different pouring targets (mainly due to the measurement delays). 
% % This means the relative error reduces as the target amount increases.
% % \textcolor{blue}{On average, pouring takes  \textcolor{blue}{X} sec to perform.}
% In Fig.~\ref{fig:pour_control}, $e(t)$ stays fixed in the pre-pouring phase, then decreases as the pouring proceeds (the end-effector velocity for pouring decreased accordingly).
% In the post-pouring phase, the robot end-effector returns to its home configuration.
% At the beginning of this last phase, the measurements still change due to delayed sensor outputs.

% \paragraph*{\textbf{Pouring}} The pouring skill evaluation is shown in Fig.~\ref{fig:pouring}.% using water as the solvent.
% The accuracy of pouring (Fig.~\ref{fig:pour_violin}) is measured by the \textit{difference} between the target and the actual poured amount.
% % The result of accuracy evaluation is shown in Fig.~\ref{fig:pour_violin}.
% The absolute error is around 2 g for different pouring targets (mainly due to the measurement delays), which means the relative error reduces as the target amount increases.
% % While in the case of target weight 10g undershoots, in other cases it overshoots.
% \textcolor{blue}{In average, pouring takes X seconds to perform.}
% % These errors are mainly caused by the scale delayed feedbacks by $\sim$3s.
% % An example of pouring control input and output signals is shown in Fig.~\ref{fig:pour_control}.
% % The end-effector is rotated back and forth by the shaping function.
% In Fig.~\ref{fig:pour_control}, the difference between the target amount and weight feedback stays fixed in pre-pouring phase, it decreases as the pouring proceeds (the end-effector velocity for pouring decreased accordingly), and finally the robot end-effector goes back to home configuration in post-pouring phase. Notice that, at the beginning of post-pouring phase, while not pouring, the error reduces due to the delayed weight measurements.
% % The delay of weight feedback is shown as the difference between end of pouring and stabilization of weight.
% % Although there is a small overshoot in this experiment, we attain a reasonable accuracy.

\subsection{Solubility Experiments}
\label{experiment-solubility}
\begin{figure*}[htb]
\begin{minipage}{0.75\linewidth}
    \includegraphics[width=0.9\linewidth]{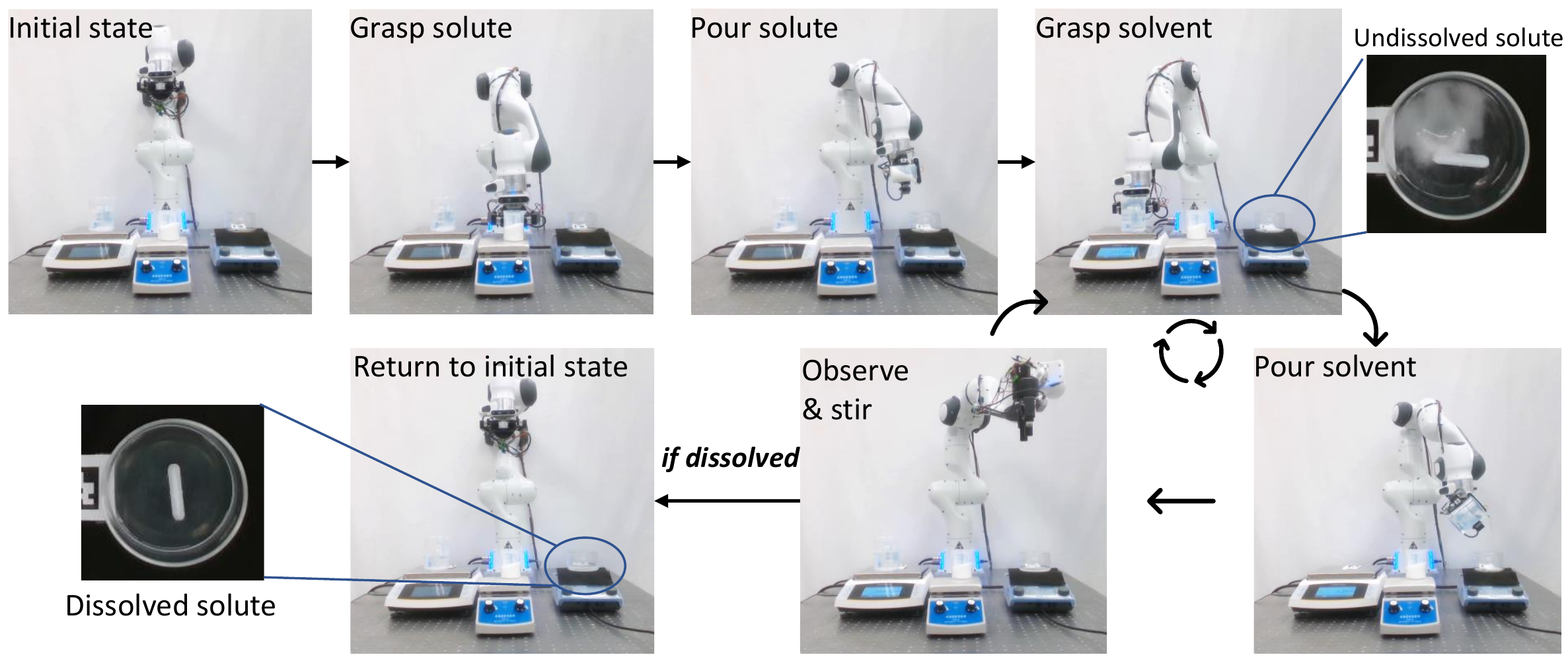}
    \centering
    \end{minipage}
    \begin{minipage}{0.23\linewidth}
    \caption{\textbf{Workflow of solubility experiment.}
    First, a fixed amount of the solute is added to the dish on the weighing scale and stirrer.
    The robot pours 10 g of water into the dish.
    The solution is mixed with the magnetic stirrer.
    After stirring, the turbidity of the solution is measured to check dissolvement.
    If undissolved, another 10 g of water is added until no solutes remain.
    The experiment was conducted at room temperature (25$^\circ$C).
    \vspace{0cm}
    }
    \label{fig:solubility}
\end{minipage}

\end{figure*}

We measured the solubility of three solutes, table salt (sodium chloride), sugar (sucrose), and alum (aluminum potassium sulfate).
The workflow of the solubility experiments is shown in Fig.~\ref{fig:solubility}.
The robot continues pouring water until turbidity stops changing.
The measured solubility for three solutes is shown in Table~\ref{tab:solubility}.
The robot framework managed to measure the solubility with accuracy that is comparable to solubility values found in the literature~\cite{nationalhandbook}.

% We first add a fixed amount of the solute to the bowl on the weighing scale and stirrer.
% The robot poured a small amount of water into the beaker.
% We chose initial amount based on the expected solubility of each solute from the literature.
% A stir bar rotated by the magnetic stirrer continuously mixed the solution.
% After pouring, we waited three minutes and visually checked whether all solutes are dissolved.
% If solutes are remained, the robot added another 5 g of water.
% If all solutes dissolved into water, we measured the total amount of poured water by the difference of initial and final weight of beaker, and calculated the solubility.
% The experiment was conducted in a room temperature (21 $^\circ$C).

\vspace{-0.6em}
\begin{table}[ht]
\centering
\caption{\textbf{Results of the solubility experiments.} Amount of solute in the beaker, amount of water to dissolve all solute, calculated solubility (the amount of solute dissolved per 100 g of water), and literature data for solubility at 20$^\circ$C is shown. Literature data are taken or calculated from \cite{nationalhandbook}.
}
\setlength{\tabcolsep}{1pt}

\label{tab:solubility}
\begin{tabular}{@{} C{1.2cm}C{1.30cm}C{1.30cm}C{1.30cm}C{1.30cm}C{1.30cm} @{}} \toprule
    \textbf{\textit{solute}} & \textbf{\textit{solute [g]}} & \textbf{\textit{water [g]}} & \textbf{\textit{solubility}} & \textbf{\textit{lit. data}}& \textbf{\textit{\% error}}\\ \hline
 %\rowcolor[rgb]{0.906,0.902,0.902}    Salt & 8.7 & 31.22 & 27.9 & 35.8  & 22.1 \\ \hline % result of the second experiment
 \rowcolor[rgb]{0.906,0.902,0.902}    Salt & 13.9 & 41.8 & 33.2 & 35.8  & 7.2 \\ \hline % if we use the first experiment
     Sugar & 60.00 & 26.46 & 226.8 & 203.9 & 11.2 \\ \hline
 \rowcolor[rgb]{0.906,0.902,0.902}    Alum & 3.00 & 29.87 & 10.0 & 11.4 & 12.3\\ \bottomrule
\end{tabular}
\end{table}

\begin{figure}[htb]
    \includegraphics[width=0.8\linewidth]{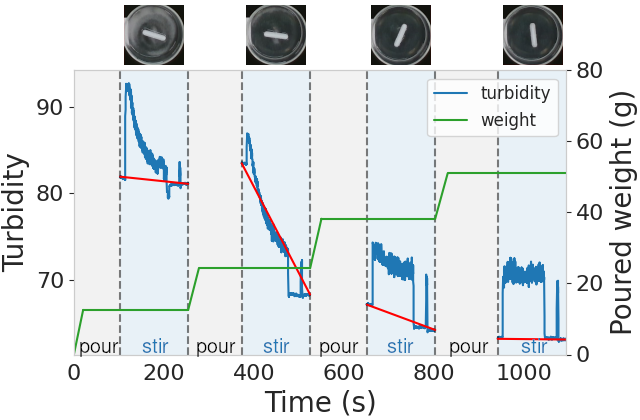}
    \centering
    \caption{\textbf{Turbidity change during experiment.} Water is poured into the dish during pouring (grey) and turbidity is measured during observation (blue). The end of the experiment is determined by turbidity comparison. In this example, all solutes are dissolved at the third pouring because the turbidity change after the fourth pouring is below the threshold. The average weight of the second and third pouring is used to calculate the solubility.}
    \label{fig:turbidity_change}
    \vspace{-1mm}
\end{figure}

\vspace{-1em}
The primary reason for the difference from the literature value is the range of minimum amount of water required for dissolving.
% Should I put this under discussion?
In an example of turbidity change shown in Fig.~\ref{fig:turbidity_change}, the robot can only tell the second pouring is insufficient and the third pouring is sufficient to dissolve all solutes, but it cannot tell the exact required amount.
As a result, the solubility measurement inherently includes error caused by the resolution of pouring.
We can reduce the error by pouring a smaller amount of water at once, but pouring smaller than 10 g is difficult because of the delayed feedback of the scale and the scale minimum resolution.
We can improve the accuracy of solubility measurement by developing a pipette designed for a robot.

We evaluated the success rate of the solubility experiment by conducting several trials.
An experiment is considered successful when the robot could execute all actions without aborting and could calculate solubility.
%An experiment is aborted when the robot arm or in-hand beaker collides with other objects, or a notable amount of water is spilled.
%An experiment is considered successful when solubility could be measured.
Without execution time refinement, 5 out of 9 experiments were successful, i.e., $\sim56\%$ success rate. Further results are available on the website.
%Note that we excluded human errors, such as misplacing glassware with respect to markers or incorrectly running software or hardware, to measure the accuracy of the proposed framework.
Note that human errors, such as incorrect software handling, were excluded from the calculation.
%The first mode of failure was the \textit{inconsistency between the simulated object models and their physical counterparts}, leading to collisions during execution.
%Secondly, \textit{perception errors} for object poses caused paths that were valid in simulation to collide with physical objects during execution.
%Lastly, the \textit{TAMP module failed} to find a feasible plan for two experiments due to infeasible pose goal requests for constrained inverse kinematics or motion planning.
Two types of failures were observed; i) TAMP failure: a plan was not found because of infeasible inverse kinematics or motion planning request; ii) inconsistency between planning and actual environment: perception error and simplification of object shape in planner caused inconsistency of two environments and collisions occurred at execution time.

% \iffalse
% \begin{table}[!t]
% % \centering
% \caption{\textbf{Results of the proposed planning pipeline on different number of steps needed to complete the experiment.}
% While the proposed constrained TAMP satisfies the motion constraints, the planning success rate and time are not affected with respect to the unconstrained motion plan.}
% % reduce the column indentation
% \setlength\tabcolsep{1.0pt} % default value: 6pt
% \begin{center}
% \label{tab:result_tamp}
% \begin{tabular}{@{} C{1cm}C{1.0cm}C{1.8cm}C{0.25cm}C{1.0cm}C{1.8cm}C{1cm} @{}}
% \toprule
% \multirow{2}{*}{{\textit{\textbf{steps}}}} & \multicolumn{2}{c}{\textit{\textbf{constrained}}} &  & \multicolumn{2}{c}{\textit{\textbf{unconstrained}}} &{{\textit{\textbf{steps}}}} \\
% \cline{2-3}\cline{5-6}
%     & planning success & time [s] & & planning success & time [s] \\
% \hline
% %\\           
% \rowcolor[rgb]{0.906,0.902,0.902} 3       & 100\% & $(83.0\pm.2)$   &  & 100\% & $(82.90\pm.03)$  \\
%  5 & 75\% & $(134.72\pm.06)$ & & 100\% & $(134.68\pm.02)$\\
%  \rowcolor[rgb]{0.906,0.902,0.902} 7 & - & - & & - & -\\
% \bottomrule
% \end{tabular}
% \end{center}
% \end{table}

% \fi

\subsection{Recrystallization Experiments}
\label{experiment-recrystallization}
\begin{figure}[ht]
\begin{minipage}{0.21\linewidth}
\centering
    \includegraphics[width=0.9\linewidth, trim={7cm 0 7cm 0}, clip]{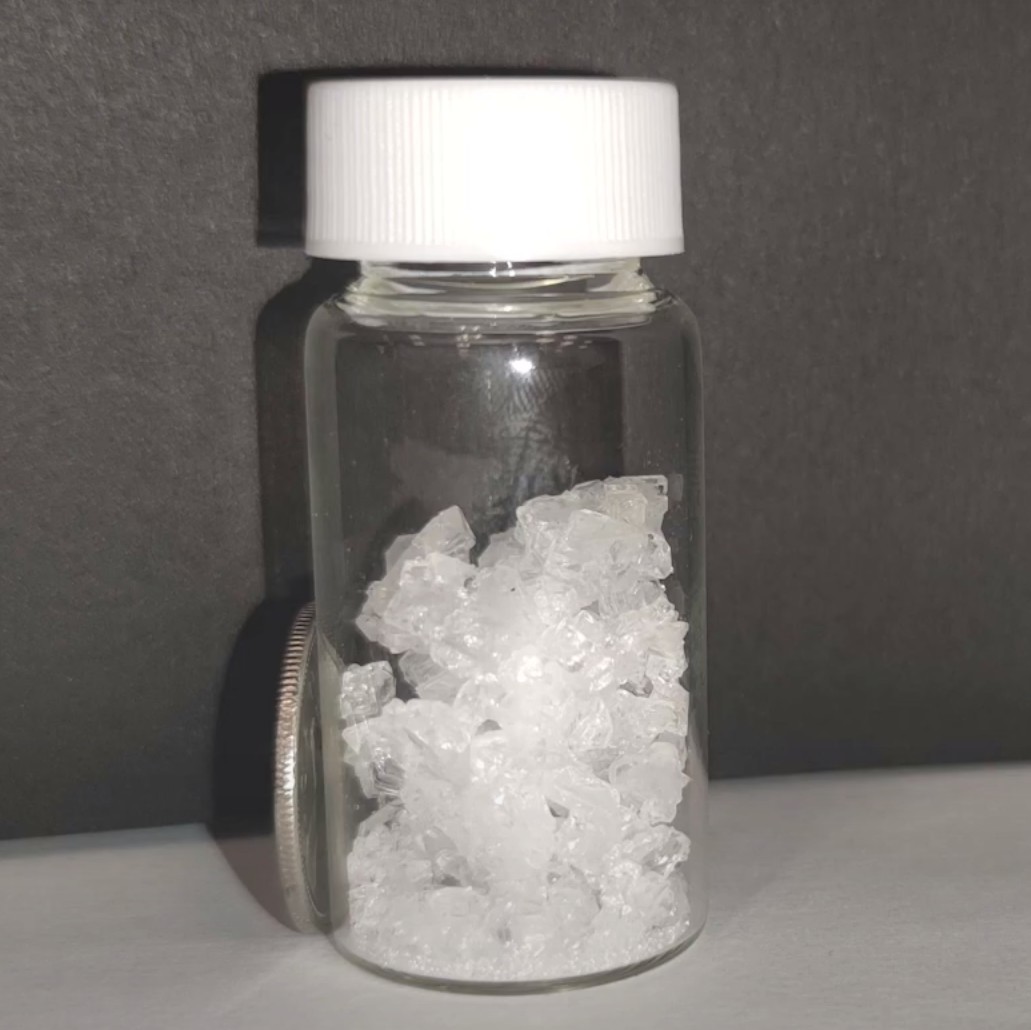}
\end{minipage}
\begin{minipage}{0.75\linewidth}
    \centering
    \caption{\textbf{Recrystallization of alum inside the water.}
    % After placing the beaker with water on top of a hotplate, it pours a beaker containing alum dish on the weighing scale.
    % The robot poured hot water into the dish with alum. Later, the solution was simultaneously heated by the hotplate and was stirred by the magnetic stirrer.
    % When all the solute dissolved, the dish was cooled down by changing the temperature of the hotplate.
    % After the solution cooled to room temperature, we observed the formation of precipitate.
    % % Precipitation of alum was observed after cooling the solution.
    After heating water by putting a beaker with water on a hotplate, the robot poured alum into a dish. The robot then poured hot water, and the solution was heated and stirred.
    The formation of a precipitate is observed after the dish is cooled down.
    The dried crystals in a vial are shown.
    %The recrystallization experiment setup extends the solubility test by pre-heating the solvent, i.e., the robot places the beaker with water on top of a hotplate. Then, a beaker containing alum was poured into the beaker on top of the weighing scale.
    %The robot poured hot water into the beaker with alum. Later, the beaker was simultaneously heated by the hotplate, and the solution was stirred by the magnetic stirrer.
    %When all the solute dissolved, the robot started to cool down the beaker by changing the temperature of the hotplate.
    %After waiting X minutes, the precipitation was observed.
    }
        \label{fig:recrystallization}
        \vspace{-0.3cm}
\end{minipage}
\end{figure}

% As an extension to the solubility experiment, we conducted a recrystallization experiment.
Recrystallization is a purifying technique to obtain crystals of a solute by using the difference in solubility at different temperatures.
Typically, solutes have higher solubility at high temperatures, meaning hot solvents will dissolve more solute than cool solvents. The excess amount of solute that cannot be dissolved anymore while cooling the solvent precipitates and forms crystals.
We tested the recrystallization of alum by changing the temperature of the water.
Alum was chosen as the target solute since its solubility greatly changes according to water temperature.
The recrystallization experiment setup extends the solubility test by pre-heating the solvent.
Fig.~\ref{fig:recrystallization} show the result of the experiment.

% The recrystallization experiment setup extends the solubility test by pre-heating the solvent, i.e., the robot places the beaker with water on top of a hotplate. Then, a beaker containing alum was poured into the beaker on top of the weighing scale.
% The robot poured hot water into the beaker with alum. Later, the beaker was simultaneously heated by the hotplate, and the solution was stirred by the magnetic stirrer.
% When all the solute dissolved, the robot started to cool down the beaker by changing the temperature of the hotplate.
% After the solution cooled to the room temperature, we observed the formation of precipitate.
% Fig.~\ref{fig:recrystallization} show the result of the experiment.
%Fig.~\ref{fig:recrystallization} demonstrates the recrystallization experiment done by the robot.
%This process can be applied to the purification of a substance and is known as a basic operation in chemistry.

% We tested the recrystallization of alum by changing the temperature of the water.
% Alum was chosen as a target since the solubility greatly changes according to water temperature.
% The experiment setup is similar to the solubility test.
% The beaker containing alum was put on top of the weighing scale.
% The robot poured water into the beaker, the beaker was simultaneously heated by the hotplate, and the solution was stirred by the magnetic stirrer.
% When all of the solutes dissolved, we started to cool down the beaker by changing the temperature of the hotplate.
% After waiting X minutes, the precipitation was observed, as shown in Figure Z.

\vspace{-3pt}
\subsection{Discussion and Limitations}
Our results show the proposed TAMP pipeline can effectively perform multi-step chemistry experiments from an experiment description in XDL.
However, the current study is limited to two types of chemistry experiments because the number of skills incorporated in the framework is limited.
Increasing the repertoire of skills, such as glassware perception in 3D and clutter, without fiducial markers~\cite{eppel2020computer}, can improve the framework scalability. % into more diverse experiments.
% In terms of perception, the current use of fiducial markers inhibits the application of the proposed framework in a more realistic environment.
% A visual perception system that detects glassware, estimates their poses, and identifies vessel content and label will allow deploying the framework to the real environments~\cite{eppel2020computer}.
% Although the proposed pouring algorithms showed promising results, manual parameter tuning is required for the best performance. %may be limited to low-viscosity fluids and standard vessels.
% Hence, learning chemistry manipulation techniques is an important direction for future research.
%further development is necessary to increase the utility of the framework.
% without significant additional computational cost.
%However, orientation constraints limit the robot's operational workspace.
Also, PDDLStream inefficiency inhibits the framework from being reactive in a dynamic environment.
%However, it is not possible to run TAMP online.
%This limits how quickly the system can react and replan in a dynamic environment, which can become problematic in the presence of humans. 
%\textcolor{blue}{Recently, learning-based search heuristics for PDDLStream has been proposed to enhance its efficiency~\cite{khodeir2021learning}.}
Incorporating the learning-based search heuristics for PDDLStream~\cite{khodeir2021learning} may overcome this limitation.
%Table~\ref{tab:constrained-motion-plan} and chemistry experiment results demonstrate that 
Constrained motion planning was shown to effectively avoid spillage of the beaker contents during transfer in our experiments. We have also shown that adding an extra 8th DoF to the robot enabled more flexibility and a higher success rate for constrained motion planning. % and consequently chemistry experiments.
However, the proposed constrained motion planning embedded in TAMP cannot run in real-time. Considering the dynamics of the beaker content may help to have higher flexibility in robot manipulation~\cite{muchacho2022solution}.
%Improving the success rate of constrained motion planning is necessary to increase the overall success rate of planning and execution.
Although our skill currently has attained 8\% error for liquid and powder pouring, higher accuracy is desirable for precise experiments. %in a chemistry lab.
We used a scale with integrated functionality for stirring and heating, but its measurement is delayed for 3 s.
Higher precision pouring can be attained using a scale with a shorter response time; also, it can be achieved by specialized tools, such as a pipette.
In addition, visual feedback during pouring may avoid spillage.%may avoid spillage and improve accuracy.

\section{Conclusion}
\label{sec:conclusions}
In this paper, we introduced a chemistry lab automation framework using general-purpose robot manipulators that execute long-horizon chemistry experiments in a closed-loop fashion, being able to visually inspect the completion of the experiment. The approach adapts to different chemistry experiments by transforming high-level experiment descriptions written in XDL into a sequence of subgoals planned by TAMP solver. We incorporated perception, constrained TAMP, an 8-DoF robot, and interaction with heterogeneous lab tools and equipment interfaces into our framework. We demonstrated the capabilities of our system by performing different pouring skills, which enabled the robot to handle solubility and recrystallization experiments. % in semi-structured workspaces found in existing labs.
%executing constraint-satisfying plans, 
%Through incorporating perception, constrained task and motion planning with PDDLStream, 7+1 DoF robot, and pouring and lab tools skill interfaces, the framework could handle experiments in semi-structured workspaces found in existing labs.
%interfacing with chemistry lab hardware, 
%and completing. % while monitoring their progress in a closed-loop fashion.
We aim to accelerate materials discovery by extending this framework to perform more reliable and effective experiments by learning various skills using simulation tools. We would like to enhance the chemists' experience by introducing a natural language interface.
% In this paper, we introduced a framework toward automated chemical experiments using robot manipulators.
% The approach adapts to different experiments by leveraging widely-used chemistry experiment descriptions written in XDL and existing lab tools.
% Through incorporating perception, constrained task and motion planning with PDDLStream, 7+1 DoF robot, and pouring and lab tools skill interfaces, the framework could handle experiments in semi-structured workspaces found in existing labs.
% We demonstrated the capabilities of our framework through developing a system that performs three different pouring skills, executes constraint-satisfying plans, interfaces with chemistry lab hardware, and completes solubility measurement and recrystallization experiments while monitoring their progress in a closed-loop fashion. We aim to accelerate materials discovery by extending this framework to perform more reliable and effective experiments by learning new skills leveraging simulation tools.

%%%%%%%%%%%%%%%%%%%%%%%%%%%%%%%%%%%%%%%%%%%%%%%%%%%%%%%%%%%%%%%%%%%%%%%%%%%%%%%%

% \clearpage
% Having error here
%\renewcommand*{\bibfont}{\small}
\bibliographystyle{IEEEtran}
\bibliography{reference}

\begin{thebibliography}{10}
\providecommand{\url}[1]{#1}
\csname url@rmstyle\endcsname
\providecommand{\newblock}{\relax}
\providecommand{\bibinfo}[2]{#2}
\providecommand\BIBentrySTDinterwordspacing{\spaceskip=0pt\relax}
\providecommand\BIBentryALTinterwordstretchfactor{4}
\providecommand\BIBentryALTinterwordspacing{\spaceskip=\fontdimen2\font plus
\BIBentryALTinterwordstretchfactor\fontdimen3\font minus
  \fontdimen4\font\relax}
\providecommand\BIBforeignlanguage[2]{{%
\expandafter\ifx\csname l@#1\endcsname\relax
\typeout{** WARNING: IEEEtran.bst: No hyphenation pattern has been}%
\typeout{** loaded for the language `#1'. Using the pattern for}%
\typeout{** the default language instead.}%
\else
\language=\csname l@#1\endcsname
\fi
#2}}

\bibitem{seifrid2022autonomous}
M.~Seifrid, \emph{et~al.}, ``Autonomous chemical experiments: Challenges and
  perspectives on establishing a self-driving lab,'' \emph{Acc. Chem. Res.},
  vol.~55, no.~17, pp. 2454--2466, 2022.

\bibitem{abolhasani2023rise}
M.~Abolhasani and E.~Kumacheva, ``The rise of self-driving labs in chemical and
  materials sciences,'' \emph{Nat. Synth.}, pp. 1--10, 2023.

\bibitem{menard2020review}
A.~D. M{\'e}nard and J.~F. Trant, ``A review and critique of academic lab
  safety research,'' \emph{Nature chemistry}, vol.~12, no.~1, pp. 17--25, 2020.

\bibitem{xu2021seeing}
H.~Xu, \emph{et~al.}, ``Seeing glass: Joint point-cloud and depth completion
  for transparent objects,'' in \emph{Ann. Conf. on Robot Learning}, 2021.

\bibitem{Wang2023MVTrans}
Y.~R. Wang, \emph{et~al.}, ``Mvtrans: Multi-view perception of transparent
  objects,'' \emph{arXiv:2302.11683}, 2023.

\bibitem{shiri2021automated}
P.~Shiri, \emph{et~al.}, ``Automated solubility screening platform using
  computer vision,'' \emph{Iscience}, vol.~24, no.~3, p. 102176, 2021.

\bibitem{fakhruldeen2022archemist}
H.~Fakhruldeen, \emph{et~al.}, ``{ARChemist}: Autonomous robotic chemistry
  system architecture,'' \emph{arXiv:2204.13571}, 2022.

\bibitem{ICAPS20paper186}
C.~R. Garrett, \emph{et~al.}, ``{PDDLStream}: Integrating symbolic planners and
  blackbox samplers via optimistic adaptive planning,'' in \emph{Proceedings of
  the 30th Int. Conf. on Automated Planning and Scheduling ({ICAPS})}.\hskip
  1em plus 0.5em minus 0.4em\relax {AAAI} Press, 2020, pp. 440--448.

\bibitem{Kennedy2019Autonomous}
M.~Kennedy, \emph{et~al.}, ``Autonomous precision pouring from unknown
  containers,'' \emph{IEEE Robo. and Automation Letters}, vol.~4, no.~3, p.
  2317–2324, Jul 2019.

\bibitem{Huang2021Robot}
Y.~Huang, \emph{et~al.}, ``\BIBforeignlanguage{en}{Robot gaining accurate
  pouring skills through self-supervised learning and generalization},''
  \emph{\BIBforeignlanguage{en}{Robo. and Autonomous Systems}}, vol. 136, p.
  103692, Feb 2021.

\bibitem{burger2020mobile}
B.~Burger, \emph{et~al.}, ``A mobile robotic chemist,'' \emph{Nature}, vol.
  583, no. 7815, pp. 237--241, 2020.

\bibitem{mehr2020universal}
S.~H.~M. Mehr, \emph{et~al.}, ``A universal system for digitization and
  automatic execution of the chemical synthesis literature,'' \emph{Science},
  vol. 370, no. 6512, pp. 101--108, 2020.

\bibitem{aeronautiques1998pddl}
C.~Aeronautiques, \emph{et~al.}, ``{PDDL} - the planning domain definition
  language,'' \emph{Tech. Rep.}, 1998.

\bibitem{doi:10.1177/0278364910396389}
D.~Berenson, \emph{et~al.}, ``Task space regions: A framework for
  pose-constrained manipulation planning,'' \emph{Int. J. Robot. Res.},
  vol.~30, no.~12, pp. 1435--1460, 2011.

\bibitem{muchacho2022solution}
R.~I.~C. Muchacho, \emph{et~al.}, ``A solution to slosh-free robot trajectory
  optimization,'' in \emph{2022 IEEE/RSJ Int. Conf. on Intelligent Robots and
  Systems (IROS)}.\hskip 1em plus 0.5em minus 0.4em\relax IEEE, 2022, pp.
  223--230.

\bibitem{Kingston2019Exploring}
Z.~Kingston, \emph{et~al.}, ``Exploring implicit spaces for constrained
  sampling-based planning,'' \emph{Int. J. Robot. Res.}, vol.~38, no. 10-11,
  pp. 1151--1178, 2019.

\bibitem{kitchener2017review}
B.~G. Kitchener, \emph{et~al.}, ``A review of the principles of turbidity
  measurement,'' \emph{Progress in Physical Geography}, vol.~41, no.~5, pp.
  620--642, 2017.

\bibitem{olson2011apriltag}
E.~Olson, ``Apriltag: A robust and flexible visual fiducial system,'' in
  \emph{2011 IEEE Int. Conf. on Robo. and automation}, 2011.

\bibitem{helmert2006fast}
M.~Helmert, ``The fast downward planning system,'' \emph{J. Artif. Intell.
  Res.}, vol.~26, pp. 191--246, 2006.

\bibitem{Beeson_Ames_2015}
P.~Beeson and B.~Ames, ``{TRAC-IK}: An open-source library for improved solving
  of generic inverse kinematics,'' in \emph{2015 IEEE-RAS 15th Int. Conf. on
  Humanoid Robots (Humanoids)}, Nov 2015.

\bibitem{karaman2011sampling}
S.~Karaman and E.~Frazzoli, ``Sampling-based algorithms for optimal motion
  planning,'' \emph{Int. J. Robot. Res.}, vol.~30, no.~7, pp. 846--894, 2011.

\bibitem{kingston2018sampling}
Z.~Kingston, \emph{et~al.}, ``Sampling-based methods for motion planning with
  constraints,'' \emph{Annu. Rev. Control Robot. Auton. Syst.}, vol.~1, pp.
  159--185, 2018.

\bibitem{kavraki1996probabilistic}
L.~E. Kavraki, \emph{et~al.}, ``Probabilistic roadmaps for path planning in
  high-dimensional configuration spaces,'' \emph{IEEE Trans Rob Autom.},
  vol.~12, no.~4, pp. 566--580, 1996.

\bibitem{doi:10.1177/027836498500400201}
T.~Yoshikawa, ``Manipulability of robotic mechanisms,'' \emph{Int. J. Robot.
  Res.}, vol.~4, no.~2, pp. 3--9, 1985.

\bibitem{zhang2020modular}
K.~Zhang, \emph{et~al.}, ``A modular robotic arm control stack for research:
  {Franka-Interface} and {FrankaPy},'' \emph{arXiv:2011.02398}, 2020.

\bibitem{coleman2014reducing}
D.~Coleman, \emph{et~al.}, ``Reducing the barrier to entry of complex robotic
  software: a {MoveIt!} case study,'' \emph{arXiv:1404.3785}, 2014.

\bibitem{elion}
``elion: Constrained planning in {MoveIt} using {OMPL}'s constrained planning
  interface,'' \url{https://github.com/JeroenDM/elion}, 2020.

\bibitem{wolthuis1960determination}
E.~Wolthuis, \emph{et~al.}, ``Determination of solubility: a laboratory
  experiment,'' \emph{J. of Chemical Education}, vol.~37, no.~3, p. 137, 1960.

\bibitem{nationalhandbook}
N.~A.~O. of~Japan, \emph{Handbook of Scientific Tables}.\hskip 1em plus 0.5em
  minus 0.4em\relax WORLD SCIENTIFIC, 2022.

\bibitem{eppel2020computer}
S.~Eppel, \emph{et~al.}, ``Computer vision for recognition of materials and
  vessels in chemistry lab settings and the vector-labpics data set,''
  \emph{ACS central science}, vol.~6, no.~10, pp. 1743--1752, 2020.

\bibitem{khodeir2021learning}
M.~Khodeir, \emph{et~al.}, ``Learning to search in task and motion planning
  with streams,'' \emph{arXiv:2111.13144}, 2021.

\end{thebibliography}

% \clearpage
%\appendix
%\import{sections/}{appendix.tex}

\end{document}